\documentclass[journal,twoside,web]{ieeecolor}
\usepackage[table]{xcolor}
\usepackage{tmi}
\usepackage{cite}
\usepackage{amsmath,amssymb,amsfonts}
\usepackage{algorithmic}
\usepackage{graphicx}
\usepackage{textcomp}
\usepackage{multirow}
\usepackage{booktabs}
\usepackage{makecell}
\usepackage{svg}
\usepackage{algorithm}
\usepackage{bm} 

\usepackage{etoolbox}
\makeatletter
\@ifundefined{color@begingroup}%
{\let\color@begingroup\relax
\let\color@endgroup\relax}{}%
\def\fix@ieeecolor@hbox#1{%
\hbox{\color@begingroup#1\color@endgroup}}
\patchcmd\@makecaption{\hbox}{\fix@ieeecolor@hbox}{}{\FAILED}
\patchcmd\@makecaption{\hbox}{\fix@ieeecolor@hbox}{}{\FAILED}

\def\BibTeX{{\rm B\kern-.05em{\sc i\kern-.025em b}\kern-.08em
    T\kern-.1667em\lower.7ex\hbox{E}\kern-.125emX}}
\begin{document}
\title{Expert-Like Reparameterization of Heterogeneous Pyramid Receptive Fields in Efficient CNNs for Fair Medical Image Classification}

\author{Xiao Wu, Xiaoqing Zhang, Zunjie Xiao, Lingxi Hu, Risa Higashita, Jiang Liu,~\IEEEmembership{Senior~Member,~IEEE}
\thanks{This work is supported in part by the National Key R\&D Program of China (No.2024YFE0198100), the National Natural Science Foundation of China (No.82272086), and the Shenzhen Medical Research Fund (No.D2402014). (Xiao Wu and Xiaoqing~Zhang contribute equally.)(Corresponding author: Risa Higashita and Jiang Liu.)}
\thanks{Xiao Wu, Xiaoqing~Zhang, Zunjie Xiao, Lingxi Hu, Risa Higashita, and Jiang Liu are with Research Institute of Trustworthy Autonomous Systems and Department of Computer Science and Engineering, Southern University of Science and Technology, Shenzhen, 518055, China. (e-mail: 12332436@mail.sustech.edu.cn, xq.zhang2@siat.ac.cn, 11930387@mail.sustech.edu.cn, lxh246@student.bham.ac.uk, risa@mail.sustech.edu.cn, liuj@sustech.edu.cn.)}
\thanks{Xiaoqing~Zhang is with Center for High Performance Computing and Shenzhen Key Laboratory of Intelligent Bioinformatics, Shenzhen Institute of Advanced Technology, Chinese Academy of Sciences, Shenzhen, 518055, China.}
\thanks{Risa Higashita is also with Tomey Corporation, Nagoya, 451-0051, Japan; Changchun University, Changchun, 130022, China.}
\thanks{Jiang Liu is also with School of computer science, University of Nottingham Ningbo China, Ningbo 315100, China;  School of Ophthalmology and Optometry, Wenzhou Medical University, Wenzhou 325035, China; Changchun University, Changchun, 130022, China}}
\maketitle

\begin{abstract}
Efficient convolutional neural network (CNN) architecture design has attracted growing research interests. However, they typically apply single receptive field (RF), small asymmetric RFs, or pyramid RFs to learn different feature representations, still encountering two significant challenges in medical image classification tasks: \textit{\romannumeral1)} They have limitations in capturing diverse lesion characteristics efficiently, e.g., tiny, coordination, small and salient, which have unique roles on the classification results, especially imbalanced medical image classification. \textit{\romannumeral2)} The predictions generated by those CNNs are often unfair/biased, bringing a high risk when employing them to real-world medical diagnosis conditions. To tackle these issues, we develop a new concept, Expert-Like Reparameterization of Heterogeneous Pyramid Receptive Fields (ERoHPRF), to simultaneously boost medical image classification performance and fairness. This concept aims to mimic the multi-expert consultation mode by applying the well-designed heterogeneous pyramid RF bag to capture lesion characteristics with varying significances effectively via convolution operations with multiple heterogeneous kernel sizes. Additionally, ERoHPRF introduces an expert-like structural reparameterization technique to merge its parameters with the two-stage strategy, ensuring competitive computation cost and inference speed through comparisons to a single RF. To manifest the effectiveness and generalization ability of ERoHPRF, we incorporate it into mainstream efficient CNN architectures. The extensive experiments show that our proposed ERoHPRF maintains a better trade-off than state-of-the-art methods in terms of medical image classification, fairness, and computation overhead. The code of this paper is available at https://github.com/XiaoLing12138/Expert-Like-Reparameterization-of-Heterogeneous-Pyramid-Receptive-Fields.
\end{abstract}


\begin{figure}[t]
\centering
\includegraphics[width=1.0\linewidth]{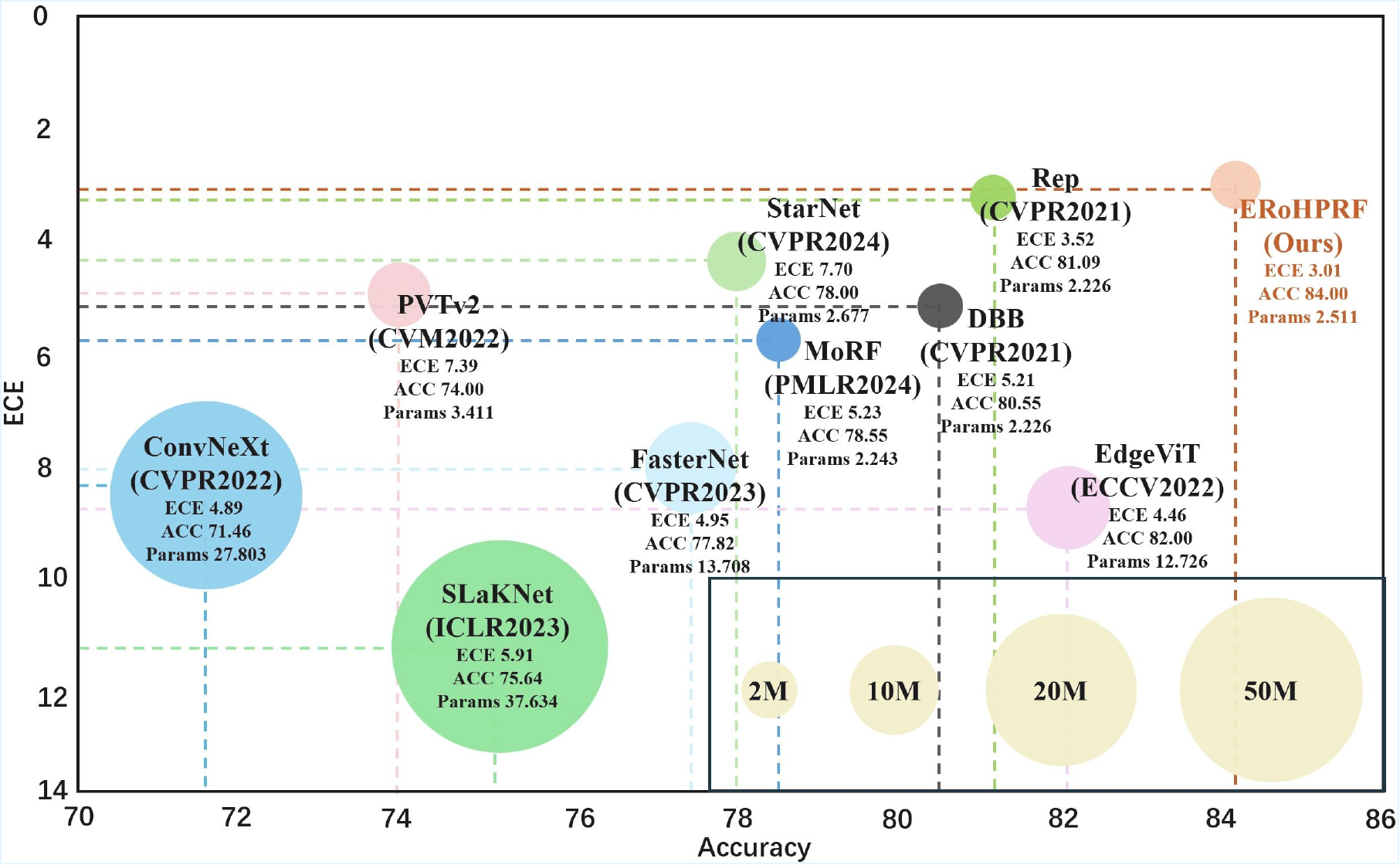}
 \caption{Comprehensive Comparisons of our ERoHPRF, existing state-of-the-art efficient DNNs, and structural reparameterization methods in terms of accuracy ($\uparrow$), expected calibration error (ECE) ($\downarrow$), and parameter number ($\downarrow$) based on the APTOS2019 dataset. The bubble size corresponds to the parameter number of all methods in the inference stage. The larger the bubble size, the greater the method's parameter number.}
   \label{InferenceCost}
\end{figure}

\section{Introduction}
\label{sec:intro}
\IEEEPARstart{C}{onvolutional} neural networks (CNNs) have become mainstream techniques in the medical image analysis field over the past years~\cite{chen2024transunet,zhang2024regional,he2020cabnet,10851813,ZHANG2022102499}, such as medical image classification, medical image segmentation, and lesion detection. A plausible reason behind their promising success is that most existing works laid emphasis on constructing complex CNN architectures by exploiting the potential of various convolution operators to improve their performance heuristically or intuitively. However, the computational resources of medical devices are often limited (e.g., wearable medical devices), and clinical diagnosis usually requires real-time feedback, bringing significant challenges to deploying these sophisticated CNNs on resource-limited medical devices.

In seeking to solve this problem, scholars have begun to design efficient CNNs and have made significant progress. MobileNet~\cite{howard2017mobilenets,zhang2024mixed,howard2019searching} pioneered the efficient CNN design by adopting the depthwise separable convolution (DSConv) operator. Then, pointwise group convolution and channel shuffle operators were developed to construct ShuffleNet \cite{2017ShuffleNet,2018ShuffleNetV2}. An interleaved group convolution operator was proposed to build IGCNet \cite{zhang2017interleaved,2018IGCNetV2}. Unfortunately, these efficient CNNs typically adopt a single receptive field (RF) (e.g., $3\times3$ and $1\times1$) in an independent convolution operator, restricting their abilities to capture diverse lesion representations for medical image classification enhancement, such as boundary, tiny, coordination, small, and salient. Recently, structural reparameterization techniques have become another mainstream paradigm, which allows efficient CNNs to maintain a good trade-off between effectiveness and efficiency via the amalgamation of diverse RFs in the inference stage. For example, the asymmetric convolution block (ACB) merges the kernel size parameters of coordination RFs into the square RF without additional parameters in the inference stage~\cite{2019ACNet}. However, these methods typically use small asymmetric RFs or pyramid RFs in an individual convolution operator, making it difficult to capture different lesion representations effectively. \textit{Here, we suggest that different lesion characteristics have unique yet important impacts on medical image classification performance, especially for imbalanced medical image classification, but these lesion representation characteristics have been under-exploited via efficient RF concept design in the individual convolution operator.}

Additionally, current CNNs often face fairness problems in medical image classification tasks \cite{luo2024fair,luo2024fairclip,park2022fair,chen2022semi}, which generate biased predictions for different diseases, sporadic diseases, due to various reasons, e.g., age, population, lesion characteristics, and imbalanced data distribution. 
This unfairness issue limits their applications to the real medical diagnosis scene. One possible reason is that they typically utilize the convolution operators with the single RF or small asymmetric RFs to capture lesion characteristics, causing them to learn specific lesion characteristics with bias and inevitably ignore the relative significance of other lesion characteristics. We suggest that convolution operators with heterogeneous RFs can compensate for the shortcomings of the existing convolution operators in addressing the fairness problem due to the inefficient diverse lesion representation capturing, but this has attracted little attention.

To address the two above-mentioned challenges, we design a novel concept: expert-like reparameterization of heterogeneous pyramid receptive fields (ERoHPRF), instead of using a single RF or small asymmetric RFs for efficient convolution operator design, which is motivated by the multi-expert consultation mode. As presented in Fig.~\ref{framework} (left lower corner), multiple experts have unique experiences and knowledge in assessing the relative significances of different lesion types in the clinical diagnosis conditions. Therefore, it is well-acknowledged that aggregating the diagnosis opinions of several experts helps make precise and fair diagnosis conclusions. To mimic this consultation mode, ERoHPRF treats each heterogeneous RF bag as an independent expert in effectively learning lesion characteristics with varying levels of significance from medical images via the well-designed heterogeneous pyramid RF bag in the convolution operator (It is worth noting that expert-like in ERoHPRF indicates heterogeneous pyramid RF bag to mimic the diagnosis process of different experts.). To aggregate different learned lesion representations and further reduce the computational overhead in the inference stage, we propose a novel expert-like structural reparameterization technique to merge the parameters of heterogeneous pyramid RF bag into one extensive RF via heterogeneous RF reparameterization and expert-like pyramid reparameterization.

To manifest the effectiveness and generalization ability of ERoHPRF in fair medical image classification, we incorporate it into mainstream efficient CNNs, such as ShuffleNet, MobileNet, and MixNet. The extensive experiments on balanced/imbalanced medical image datasets show the superiority and generalization ability of our ERoHPRF over state-of-the-art (SOTA) methods in terms of classification, fairness, and computational overhead in the inference stage. As presented in Fig.~\ref{InferenceCost}, we adopt accuracy, expected calibration error (ECE), and parameter number to indicate classification, fairness, and computational efficiency (classification and fairness metrics are significant indicators to evaluate informative representation diversity enhancement of our ERoHPRF in a lightweight manner.). In Fig.~\ref{InferenceCost}, the bubble size indicates the number of parameters; the larger the bubble size, the more parameters the method has.

The main contributions of this paper are summarized as follows:
\begin{enumerate}
  \item \textit{We are the first to provide a unique perspective to improve diverse lesion representation capturing and fairness in medical classification tasks from the aspect of heterogeneous pyramid RFs in the efficient convolution operator design.}
  \item Motivated by the multi-expert consultation mode, we propose a new concept of Expert-Like Reparameterization of Heterogeneous Pyramid Receptive Fields (ERoHPRF) for efficient CNN designing, applying heterogeneous pyramid RF bag to capture diverse lesion representations via convolution operations with multiple heterogeneous pyramid-like kernel sizes effectively. In the inference stage, we introduce an expert-like structural reparameterization technique to merge the parameters of heterogeneous pyramid RF bag to keep competitive computation cost and inference speed through comparisons to a single RF.
  \item We systematically investigate the effectiveness, efficiency, and fairness of our ERoHPRF on balanced/imbalanced medical image datasets through comparisons to existing SOTA structural reparameterization methods, efficient CNNs, and advanced Transformers.
\end{enumerate}

\section{Related Work}
\label{sec:relatedwork}

\subsection{Receptive Fields in Convolution}
The research interests of RF design in the convolution operator can be traced back to the pioneering LeNet, which utilizes $5\times 5$ RF size to tackle the task of handwritten digits~\cite{1998LeNet}. AlexNet \cite{krizhevsky2012imagenet} adopts a large RF size $11\times 11$ but increases computational cost. VGGNet~\cite{2015VGG} replaces a large RF size with a small RF size $3\times 3$ and finds that it can increase the depth of CNNs. After VGGNet, small RF size has become mainstream in various convolution operators (e.g., dilated convolution) and advanced CNN architecture design, such as GoogleNet and ResNet. Meanwhile, with the rapid development of hardware, large RF design has also regained attention from a broader perspective~\cite{dosovitskiy2021image,xiao2021early,nie2025reparameterized,huang2024wavelet,wu2023dminet}. For example, Ding et al. \cite{2022RepLKNet} extend the RF size to $31\times 31$ and achieved competitive performance. Afterward, Liu et al. \cite{2022SLaKNet} further enlarge the RF into $51\times 51$ size by following a similar idea. Loos et al.~\cite{loos2024demystifying} investigate the effects of different RFs on medical image segmentation results under U-Net. However, most previous works focused on the different role understanding of the independent RF, ignoring the complementary significances of different RFs, which is beneficial to capture diverse lesion representations. On the contrary, our ERoHPRF takes the unique effects of different RFs into consideration from the aspect of heterogeneous pyramid RFs for boosting fairness in medical image classification results.

\subsection{Structural Reparameterization Techniques}
Recently, structural reparameterization techniques have become another popular paradigm to allow efficient DNNs to maintain a good balance between computational overhead and performance in the inference stage. Ding et al.~\cite{2019ACNet} pioneers the structural reparameterization method by designing an ACB with two coordination RFs and a square RF, and then merging them into the square RF. Following this, several structural reparameterization methods have been developed to boost the general performance of CNNs by reparameterizing the corresponding RFs in a convolution operator. For example, RepLKNet~\cite{2022RepLKNet} increases the RF size to $31\times 31$ and takes it as a backbone by merging other large RF sizes. A mixture of receptive fields (MoRF)~\cite{2025receptive} merges the parameters of pyramid RFs with a routing mechanism. Unluckily, these methods ignore the relative significance of heterogeneous RFs in a bag in capturing diverse lesion representations in the medical image classification tasks from the pyramid aspect. To alleviate this problem, we design heterogeneous pyramid expert RF bags and expert-like structural reparameterization under the concept of ERoHPRF to achieve SOTA medical image classification performance and reduce computational overhead in the inference stage.

\subsection{Fairness}
Artificial intelligence (AI) fairness has been a hot yet emerging research topic in recent years, especially in the era of deep learning. The issue of AI fairness is associated with many biases, such as gender, age, race, representation types, or data distributions~\cite{zong2023medfair,jin2024fairmedfm,10944285,10871930}. In this paper, we define fairness in the medical image classification task as follows: DNNs are expected to output unbiased predictions towards diverse demographics and imbalanced medical image data. Moreover, scholars have developed several techniques to boost the AI fairness of DNNs, including loss modification, data enhancement, and representation learning. For example, Park et al. \cite{park2022fair} propose a fair supervised contrastive loss to distinguish different classes and penalize sensitive information. Lin et al. \cite{lin2023improving} introduced a marginal ranking loss to improve fairness by adopting the marginal pairwise equal opportunity. Luo et al. \cite{luo2024fairclip} change the sample distribution to better learn the fair feature representations of DNNs based on each demographic group. In contrast to previous efforts, we aim to improve the fairness of DNN-based medical image classification from a new aspect of heterogeneous pyramid RFs in the convolution operator through the efficient CNN architecture construction. Inspired by the multi-expert consultation mode, we propose a novel ERoHPRF concept for boosting fair medical image classification performance based on efficient CNNs.

\section{Methodology}
\begin{figure*}[htb]
	\centering
		\includegraphics[width=0.95\linewidth]{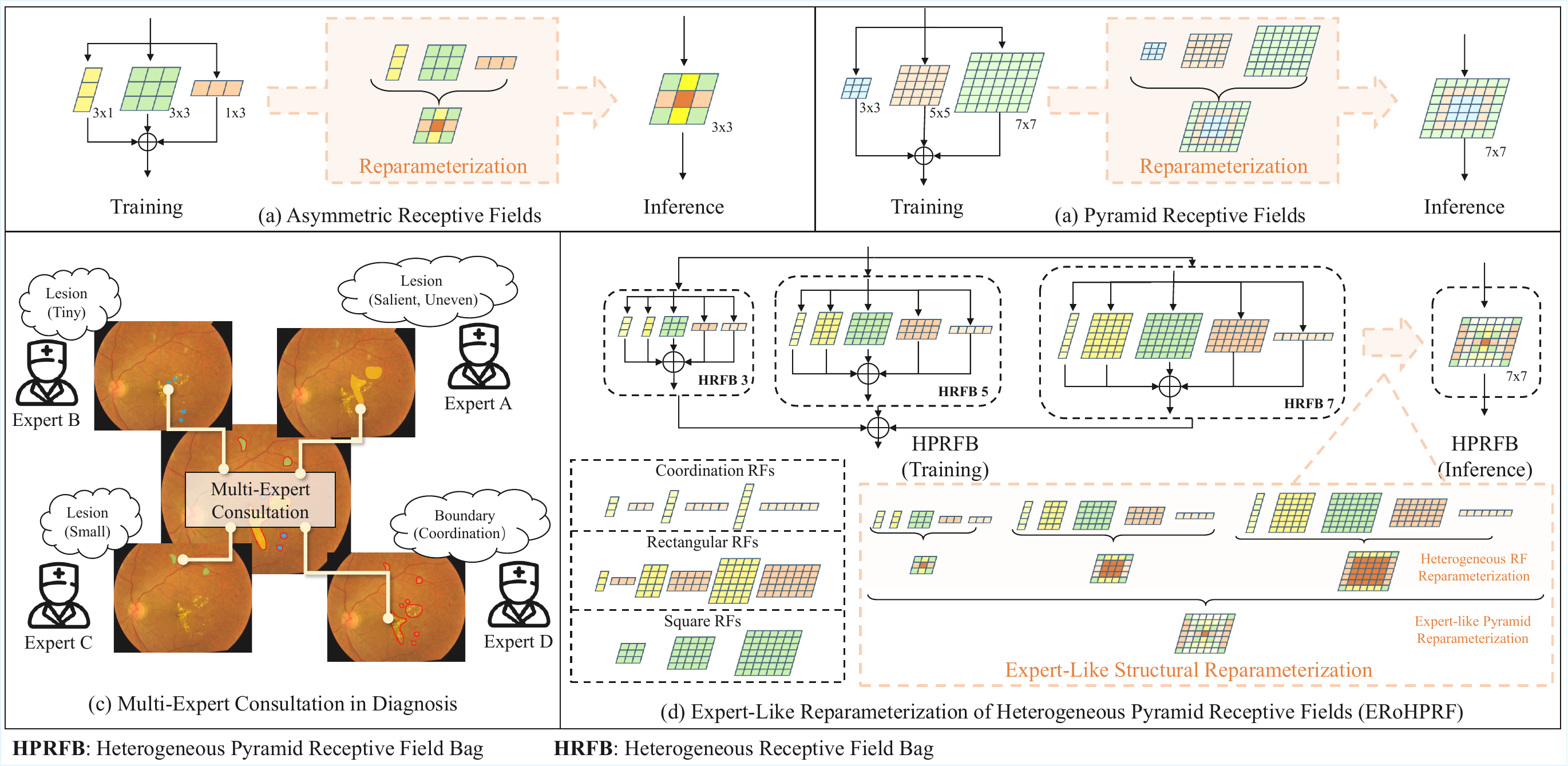}
	  \caption{(a) Asymmetric receptive fields; (b) Pyramid receptive fields; (c) Multi-expert consultation mode: each expert has unique advantages in assessing the corresponding lesions. (d) Expert-like reparameterization of pyramid receptive fields (ERoHPRF): heterogeneous pyramid receptive field bag and expert-like structural reparameterization.}
   \label{framework}
\end{figure*}

In this section, we first revisit the asymmetric and pyramid RFs and then describe our ERoHPRF in detail, as offered in Fig.~\ref{framework}.

\subsection{Revisiting Asymmetric and Pyramid RFs}
The concepts of asymmetric and pyramid RFs have been widely utilized to design different convolution methods and structural reparameterization methods, as listed in Fig.~\ref{framework}(a)-(b). Asymmetric RFs typically comprise coordination RFs and square RF (The representative implementation of it is ACB~\cite{2019ACNet}), in which the former is the indispensable complement to the latter in efficiently capturing coordination representations of single lesion characteristics, but lacks the ability to capture feature representations of diverse lesion representations simultaneously. In contrast, pyramid RFs apply multiple RF sizes to learn diverse lesion representations in a convolution operator (The representative implementation of it is MoRF~\cite{2025receptive}). Unluckily, they have limitations in obtaining coordination and boundary lesion representations. The following works of asymmetric and pyramid RFs design still cannot tackle their corresponding deficiencies.

Additionally, as shown in Fig.~\ref{framework}(c), different experts have their corresponding advantages in evaluating different lesion characteristic combinations in clinical conditions, e.g., the combination of small and salient characteristics and the combination of small and boundary characteristics. Naturally, integrating the diagnosis results of multiple experts improves diagnosis precision and fairness of diseases according to our understanding. Inspired by the multi-expert consultation mode, we design a new concept of Expert-Like Reparameterization of Heterogeneous Pyramid Receptive Fields (ERoHPRF) to capture diverse lesion representations at the same time in an individual convolution operator. Our ERoHPRF is supposed to not only take advantage of asymmetric and pyramid RFs but also overcome their shortcomings.

\subsection{Expert-Like Reparameterization of Heterogeneous Pyramid Receptive Fields}
As a convenient plug-and-play module, our ERoHPRF comprises two main components: heterogeneous pyramid RF bag (HPRFB) and expert-like structural reparameterization (ELSR). The well-designed HPRFB employ each heterogeneous receptive field bag (HRFB) with different RFs to effectively capture diverse lesion representations (e.g., tiny, small, boundary, and salient). Fig. \ref{framework}(d) presents an example of HPRFB implementation with three different RFs $\{3, 5, 7\}$ (Noting that these three RFs keep a promising balance in capturing boundary, tiny, coordination, small, and salient representations). Following it, the ELSR method applies a two-stage reparameterization strategy to merge different RFs into an expert-like reparameterization RF via heterogeneous RF bag reparameterization and expert-like pyramid parameterization for reducing computational overhead without harming the performance. Here, given the input feature map $X \in \mathbb{R}^{H\times W}$ ($H$ and $W$ indicate the height and width), we define the ERoHPRF implementation with RFs $\{3, 5, 7\}$ as follows:
\begin{equation}
Y=\begin{cases}
\phi(X,7) & \text{If Training}, \\
\hat{\phi}(X,7) & \text{Otherwise Inference},
\end{cases}
\label{eq1-1}
\end{equation}
where $Y \in \mathbb{R}^{H \times W}$ denotes the output feature map. $\phi$ and $\hat{\phi}$ represent the HPRFB operator before and after ELSR.

\paragraph{Heterogeneous Pyramid Receptive Field Bag}
The main goal of HPRFB is to capture diverse lesion representations effectively through different HRFBs with various RF types, including square RF, coordination RF, and rectangular RF, thereby boosting the classification performance and decision-making fairness of CNNs. The HPRFB consists of parallel HRFBs operator $\xi$ with pyramid RF scales, which is defined as:
\begin{equation}\
\begin{aligned}
    \phi(X,K) &=\sum_{i} \xi(X,i)\\
    &=\sum_{i}\sum_{j} \psi(X,i,j),
\end{aligned}
\label{eq2-1}
\end{equation}
where $\psi$ denotes the RF operator. $i \in \{3, 5, 7\}$ and $j \in \{\text{VC}, \text{HC}, \text{VR}, \text{HR}, \text{S}\}$ indicate the sizes and types of RFs. $\text{VC}$ and $\text{HC}$ denote vertical and horizontal coordination RFs with kernel sizes of $i \times 1$ and $1 \times i$; $\text{VR}$ and $\text{HR}$ represent vertical and horizontal rectangular RFs with kernel sizes $i \times (i{-}2)$ and $(i{-}2) \times i$; and $\text{S}$ corresponds to a square RF of kernel size $i \times i$. Specifically, the $\psi$ contains a convolution operator ($Conv$) and a batch normalization operator ($BN$). We take $\psi(X,5,\text{VR})$ as an example:
\begin{equation}
    \psi(X,5,\text{VR}) = BN(Conv_{5\times 3}(X)),
\label{eq2-2}
\end{equation}
where $Conv_{5\times 3}$ denotes the corresponding convolution operator with RF size $5\times 3$.

\paragraph{Expert-Like Structural Reparameterization} Structural reparameterization techniques have been widely utilized to merge parameters of different RFs in a convolution operator to keep a good balance between efficiency and effectiveness in the inference stage. This paper introduces an expert-like structural reparameterization (ELSR) technique to reparametrize the parameters of HPRFB through two stages: heterogeneous RF bag reparameterization and expert-like pyramid reparameterization, as shown at the bottom of Fig.~\ref{framework}(d). Before the detailed descriptions of ELSR's two stages, we first present the linearity property analysis of RFs in the convolution operator, which serves as the theoretical foundation for structural reparameterization of various RFs. Here, we take the convolution operator with a RF size of $M\times N$ as an example:
\begin{equation}
    Y' = X'\otimes W+B,
\label{eq3-1}
\end{equation}
where $\otimes$, $X'$, and $Y'$ are the convolution operator, the corresponding input and output feature maps. $W$ and $B$ are the convolution weights and biases. The output $y'_{h,w}$ at position $(h,w)$ of $Y'$ is formulated as:
\begin{equation}
y'_{h,w} = \sum_{m=1}^{M}\sum_{n=1}^{N}x'_{h-\left \lceil \frac{M}{2}  \right \rceil +m,w-\left \lceil \frac{N}{2}  \right \rceil +n}w_{m,n}+b,
\label{eq3-2}
\end{equation}
where $x'$, $w$, and $b$ are the input values, the convolution weights, and bias. From Eq.~\eqref{eq3-2}, we can easily observe the linearity of convolutions. Additionally, we discuss the homogeneity (Eq.~\eqref{eq3-3}) and additivity (Eq.~\eqref{eq3-4}) properties of the classical convolution operator as follows:
\begin{equation}
    \alpha(X'\otimes W+B) = X'\otimes (\alpha W)+(\alpha B),
\label{eq3-3}
\end{equation}
\begin{equation}
\begin{aligned}
    &X'\otimes W_{1}+B_{1}+X'\otimes W_{2}+B_{2} \\
    &= X'\otimes (W_{1}+W_{2})+(B_{1}+B_{2}),
\end{aligned}
\label{eq3-4}
\end{equation}
where $\alpha$ is a constant. $W_{1}$, $W_{2}$, $B_{1}$, and $B_{2}$ are the weights and bias of two convolution operators with the same settings, such as padding and stride.

\textbf{Heterogeneous RF Bag Reparameterization.} The main goal of this step is to merge the parameters of heterogeneous RFs in an HRFB, including the parameters of convolution and corresponding BN based on the homogeneity of convolution. According to Eq.~\eqref{eq2-2}, the BN reparameterization formula of $\psi(X,5,\text{VR})$ can be written as:
\begin{equation}
    \begin{aligned}
    \psi(X,5,\text{VR}) &= (X\otimes W_{\text{VR}}^{5} +B_{\text{VR}}^{5}-\mu)\frac{\gamma}{\sqrt{\sigma^{2}+\varepsilon}}+\beta\\
    &= X\otimes \hat{W}_{\text{VR}}^{5} +\hat{B}_{\text{VR}}^{5}
    \end{aligned},
\label{eq4-1}
\end{equation}
where $W_{\text{VR}}^{5} \in \mathbb{R}^{5\times 3}$ and $B_{\text{VR}}^{5}$ denote the convolution weights and bias. The $\mu$, $\gamma$, $\sigma$, and $\beta$ are the training parameters in the $BN$ operator, which are frozen in the inference stage. $\varepsilon$ is a tiny constant to avoid zero division. $\hat{W}_{\text{VR}}^{5}=\frac{\gamma W_{\text{VR}}^{5}}{\sqrt{\sigma^{2}+\varepsilon}}$ and $\hat{B}_{\text{VR}}^{5}=(\frac{\gamma (B_{\text{VR}}^{5}-\mu)}{\sqrt{\sigma^{2}+\varepsilon}}+\beta)$ denote the merged convolution weight and bias. Similarly, the reparameterization of other RFs, such as coordination and square, can also be implemented via Eq.~\eqref{eq4-1}. Furthermore, we use zero padding to reshape the convolution weights of heterogeneous RFs (e.g., $\text{Pad}_{5\times 3\rightarrow 5\times 5}$), we also adopt merged $\psi(X,5,\text{VR})$ as an example to explain it:
\begin{equation}
\hat{W}_{\text{VR}}^{5'} = \text{Pad}_{5\times 3\rightarrow 5\times 5}(\hat{W}_{\text{VR}}^{5}),
\label{eq4-2}
\end{equation}
where $\hat{W}_{\text{VR}}^{5'} \in \mathbb{R}^{5\times 5}$ is the padded convolution weight. Under the additivity property of convolution, we further merge heterogeneous RFs in HRFBs by taking $\xi(X,5)$ as an example:
\begin{equation}
\begin{aligned}
\xi(X,5) &= \sum_{j} \psi(X,5,j)\\
    &= X\otimes (\hat{W}_{\text{VC}}^{5'}+\hat{W}_{\text{HC}}^{5'}+\hat{W}_{\text{VR}}^{5'}+\hat{W}_{\text{HR}}^{5'}+\hat{W}_{\text{S}}^{5})\\
    &+(\hat{B}_{\text{VC}}^{5}+\hat{B}_{\text{HC}}^{5}+\hat{B}_{\text{VR}}^{5}+\hat{B}_{\text{HR}}^{5}+\hat{B}_{\text{S}}^{5})\\
    &= X\otimes \bar{W}_{\text{S}}^{5}+\bar{B}_{\text{S}}^{5}
\end{aligned},
\label{eq4-3}
\end{equation}
where $\bar{W}_{\text{S}}^{5}= (\hat{W}_{\text{VC}}^{5'}+\hat{W}_{\text{HC}}^{5'}+\hat{W}_{\text{VR}}^{5'}+\hat{W}_{\text{HR}}^{5'}+\hat{W}_{\text{S}}^{5})$ and $\bar{B}_{\text{S}}^{5}=(\hat{B}_{\text{VC}}^{5}+\hat{B}_{\text{HC}}^{5}+ \hat{B}_{\text{VR}}^{5}+\hat{B}_{\text{HR}}^{5}+\hat{B}_{\text{S}}^{5})$ are the merged convolution weight and bias. Next, this paper merges the expert-like pyramid RFs in HPRFB into an extensive RF with a single convolution operator.

\textbf{Expert-Like Pyramid Reparameterization.} This stage further fuses HRFBs with pyramid RFs of the HPRFB into an extensive one, which is expected to keep a desirable balance in classification performance, fairness, and computational complexity. First, the convolution weights of each HRFB are reshaped into a compatible size by zero padding as follows, also taking merged $\xi(X,3)$ as an example:
\begin{equation}
\bar{W}_{\text{S}}^{3\rightarrow 7} = \text{Pad}_{3\times 3\rightarrow 7\times 7}(\bar{W}_{\text{S}}^{3}),
\label{eq5-1}
\end{equation}
where $\bar{W}_{S}^{3\rightarrow 7} \in \mathbb{R}^{7\times 7}$ is the padded convolution weight. Then, we merge the weights of different HRFBs through reparameterization operation based on the additivity property of the convolution operator as follows:
\begin{equation}
\begin{aligned}
    \phi(X,7) &=\xi(X,3)+\xi(X,5)+\xi(X,7)\\
    &= X\otimes (\bar{W}_{\text{S}}^{3\rightarrow 7}+\bar{W}_{\text{S}}^{5\rightarrow 7}+\bar{W}_{\text{S}}^{7}) +(\bar{B}_{\text{S}}^{3}+\bar{B}_{\text{S}}^{5}+\bar{B}_{\text{S}}^{7})\\
    &= X\otimes \tilde{W} + \tilde{B}\\
\end{aligned},
\label{eq5-2}
\end{equation}
where $\tilde{W}= (\bar{W}_{\text{S}}^{3\rightarrow 7}+\bar{W}_{\text{S}}^{5\rightarrow 7}+\bar{W}_{\text{S}}^{7})$ and $\tilde{B}=(\bar{B}_{\text{S}}^{3}+\bar{B}_{\text{S}}^{5}+\bar{B}_{\text{S}}^{7})$ are the merged convolution weight and bias. After the two-stage structural reparameterization operations, ERoHPRF only utilizes an independent convolution operator to capture diverse lesion representations through the extensive RF, thereby achieving promising results without extra parameters and computation overhead.

Moreover, to allow audiences to understand our ERoHPRF easily, we offer its derivation analysis in the gradient backpropagation process. It is worth noting that the gradient backpropagation of ERoHPRF only occurs in the training stage, and its inference stage does not have the gradient backpropagation process. Therefore, this paper only computes the derivation of HPRFB without ELSR to explain the gradient backpropagation of ERoHPRF. According to, Eq.~\eqref{eq1-1}, Eq.~\eqref{eq2-1} and Eq.~\eqref{eq4-1}, the HPRFB in pyramid RFs of \{3, 5, 7\} can be express as:
\begin{equation}
   Y=\sum_{i}\sum_{j}(X\otimes W^{i}_{j}+B^{i}_{j}-\mu^{i}_{j})\frac{\gamma^{i}_{j}}{\sqrt{(\sigma^{i}_{j})^{2}+\varepsilon}}+\beta^{i}_{j},
   \label{eq6-1}
\end{equation}
Assuming that we adopt the classical cross-entropy (CE) loss $\mathcal{L}$ as the learning loss, we compute the derivation $\frac{\partial \mathcal{L}}{\partial X}$ of HPRFB as follows:
\begin{equation}
   \frac{\partial \mathcal{L}}{\partial X}= \frac{\partial \mathcal{L}}{\partial Y} \frac{\partial Y}{\partial X} =
   \bm{\delta} \otimes \sum_{i}\sum_{j}\frac{\gamma^{i}_{j} W^{i}_{j}}{\sqrt{(\sigma^{i}_{j})^{2}+\varepsilon}},
   \label{eq6-2}
\end{equation}
where $\bm{\delta}=\frac{\partial \mathcal{L}}{\partial Y}$ denotes the backpropagated error from the loss $\mathcal{L}$ with respect to the output $Y$.
We observe that the gradients of each RF in the HPRFB are mutually independent. Accordingly, the gradients with respect to the convolution weights and biases can be calculated as follows:
\begin{equation}
\begin{aligned}
   \frac{\partial \mathcal{L}}{\partial W_{j}^{i}} &= \frac{\partial \mathcal{L}}{\partial Y}\frac{\partial Y}{\partial W_{j}^{i}}\\
   &=\bm{\delta}\otimes\frac{\gamma^{i}_{j} X}{\sqrt{(\sigma^{i}_{j})^{2}+\varepsilon}} 
\end{aligned},
\label{eq6-3}
\end{equation}
\begin{equation}
\begin{aligned}
   \frac{\partial \mathcal{L}}{\partial B_{j}^{i}} &= \frac{\partial \mathcal{L}}{\partial Y}\frac{\partial Y}{\partial B_{j}^{i}} = \frac{\gamma^{i}_{j}}{\sqrt{(\sigma^{i}_{j})^{2}+\varepsilon}}\bm{\delta}\\
\end{aligned},
\label{eq6-4}
\end{equation}
Since the bias is a constant and broadcast during computation, Eq.~\eqref{eq6-4} can be reformulated as: 
\begin{equation}
\begin{aligned}
   \frac{\partial \mathcal{L}}{\partial B_{j}^{i}} = \frac{\gamma^{i}_{j}}{\sqrt{(\sigma^{i}_{j})^{2}+\varepsilon}}\sum\bm{\delta}.
\end{aligned}
\label{eq6-5}
\end{equation}
Based on the above analysis, ERoHPRF is capable of effectively capturing diverse lesion representations through HPRFB and corresponding helpful gradient backpropagation.

\section{Experiments}

\subsection{Datasets}
This paper utilizes three publicly available balanced/imbalanced datasets and a private imbalanced dataset to verify the effectiveness, efficiency, and fairness of ERoHPRF. These datasets involve different medical image modalities, pathologies, and patient demographics, including sensitive attributes, e.g., sex and age. The imbalanced ratios of the four datasets are listed in Table \ref{table:dataset}. The imbalance ratio denotes the ratio of samples of the most frequent class to samples of the least frequent class.

\textbf{APTOS2019 \cite{aptos2019d}.} It is a publicly available diabetic retinopathy (DR) dataset with 3,662 fundus images of five DR severity levels. This paper follows the data augmentation strategy in \cite{yang2022proco}, splitting it into three disjoint subsets: training (2,510), validation (608), and testing (544).

\textbf{ISIC2018 \cite{tschandl2018ham10000}.} It is an imbalanced dataset with 10,208 dermatoscopic images of seven different skin diseases. This study adopts the same dataset splitting and data preprocessing strategies in \cite{Zhang2023ppcr} for fair comparison.

\textbf{BTM \cite{btmd}.} It is a public brain tumor dataset with 3,264 magnetic resonance imaging (MRI) images in four tumor types under the balanced distribution: normal, meningioma, glioma, and pituitary. It is split into training (2,870) and testing (394) sets. Based on that, this study adopts 20\% of the training images as the validation set.

\textbf{AS-OCT-NC.} It is a private nuclear cataract (NC) dataset with 12,824 anterior segment optical coherence tomography (AS-OCT) images of three NC severity levels: normal, mild, and severe. It is split into three independent subsets: training (6,813), validation (2,219), and testing (3,792).

\begin{table}[h]
\caption{The details of four medical image datasets used in this paper.}
    \centering
    \resizebox{1.0\linewidth}{!}{
    \begin{tabular}{c|c|c|c|c} 
        \toprule
        Dataset&Class Number&Sample Number&Imbalance Ratio &Sensitive Attribute\\
        \midrule
        APTOS2019&5&3,662&10&-\\
        ISIC2018&7&10,208&58&Age, Sex\\
        BTM&4&3,264&2&-\\
        AS-OCT-NC&3&12,824&5&Age, Sex\\
        \bottomrule
    \end{tabular}}
    \label{table:dataset}
\end{table}

\subsection{Implementation Details}
\label{exp-setting}
\textbf{Experimental Settings.} We implement the proposed ERoHPRF, competitive structural reparameterization methods, and SOTA DNNs with the PyTorch platform and Python language. This paper runs all experiments on a server with an NVIDIA RTX 3090 GPU and an AMD EPYC 7402 CPU. The classical stochastic gradient descent (SGD) optimizer is adopted as the optimizer with default settings (a weight decay of 1e-5 and a momentum of 0.9) to update the weights of DNNs. We set the training epochs and batch size to 200 and 32, respectively. The initial learning rate (LR) is set to 0.005, decreasing by a factor of 5 every 30 epochs. Specifically, the LR is fixed at 0.000025 after 100 epochs. Standard data augmentation methods are used in the training stage, such as random flipping and cropping.

\textbf{Evaluation Metrics.} This paper utilizes classification metrics, fairness metrics, and statistics metrics to investigate the effectiveness of all methods from three different perspectives. Classification metrics include accuracy (ACC), balanced accuracy (bACC), and macro F1 (mF1) \cite{galdran2021balanced}. As for fairness metrics, we adopt the average Area Under the receiver operating characteristic Curve (AUC) for each class, expected calibration error (ECE), class-wise expected calibration error (CECE), and Brier score (BS)~ \cite{jin2024fairmedfm} as follows:
\begin{equation}
AUC = \frac{1}{S_{N}S_{P}} \sum_{i=1}^{S_{P}} \sum_{j=1}^{S_{N}}(p_{i}>p_{j}),
\end{equation}
\begin{equation}
ECE = \sum_{i=1}^{B}\frac{s_{i}}{S}|ACC(b_{i})-CON(b_{i})|,
\label{ECEF}
\end{equation}
\begin{equation}
CECE = \frac{1}{C}\sum_{c=1}^{C}ECE_{c},
\label{CECEF}
\end{equation}
\begin{equation}
BS = \frac{1}{S}\sum_{s=1}^{S} \sum_{c=1}^{C}(p_{s,c}-o_{s,c})^{2},
\end{equation}
where $S_{N}$, $S_{P}$, $S$, and $C$ indicate the number of negative, positive, all samples, and classes. $p_{i}$, $p_{j}$, and $p_{s,c}$ are the probability of the $i_{th}$ positive sample, the $j_{th}$ negative sample, and the sample $s$ for class $c$. $B$, $b_{i}$, and $s_{i}$ in Equation \eqref{ECEF} are the number of bins, the $i_{th}$ bin, and the sample in $b_{i}$ where $ACC(b_{i})$ and $CON(b_{i})$ calculate the accuracy and confidence in $b_{i}$. The indicator function $(p_{i}>p_{j})$ is 1 if $p_{i}$ is larger than $p_{j}$, and 0 otherwise. $o_{s,c}$ is 1 if sample $s$ belongs to class $c$, and 0 otherwise. Statistics metrics like p-value and 95\% confidence interval (CI) of ACC are also introduced to evaluate the performance. We also apply the number of parameters (Params), multiply-accumulate operations (MACs), and frames per second (FPS) to evaluate the model complexity/computation overhead.

\textbf{Contrast Methods.} This paper utilizes the following SOTA structural reparameterization methods to test the effectiveness of our ERoHPRF: ACB \cite{2019ACNet}, Rep \cite{2021RepVGG}, DBB \cite{2021DBB}, and MoRF \cite{2025receptive}. Moreover, we also adopt classical CNNs (ResNet~\cite{2016ResNet}, VGG~\cite{2015VGG}, and ConvNeXt~\cite{liu2022convnet}), vision transformers, multilayer perceptrons (MLP) like architectures, and efficient CNNs for comparison, such as Swin-T \cite{liu2021swin}, SLaKNet \cite{2022SLaKNet}, FasterNet \cite{chen2023run}, and StarNet \cite{ma2024rewrite}.

\begin{table*}[!htp]
    \centering
     \caption{Performance comparisons between our ERoHPRF and SOTA structural reparameterization methods in terms of medical image classification metrics, fairness metrics, and statistics metrics in the inference stage based on the APTOS2019 dataset and the ISIC2018 dataset.}
    \resizebox{1.0\linewidth}{!}{
    \begin{tabular}{c|ccc|cccc|cc|ccc|cccc|cc}  
        \toprule
         \multirow{3}{*}{Method}&\multicolumn{9}{c|}{APTOS2019}&\multicolumn{9}{c}{ISIC2018}\\
         &\multicolumn{3}{c|}{Classification Metrics}&\multicolumn{4}{c|}{Fairness Metrics}&\multicolumn{2}{c|}{Statistical significance}&\multicolumn{3}{c|}{Classification Metrics}&\multicolumn{4}{c|}{Fairness Metrics}&\multicolumn{2}{c}{Statistical significance}\\
         &ACC$\uparrow$ &bACC$\uparrow$ &mF1$\uparrow$ &AUC$\uparrow$ &ECE$\downarrow$ &BS$\downarrow$ &CECE$\downarrow$ &P-Value &95\%CI &ACC$\uparrow$ &bACC$\uparrow$ &mF1$\uparrow$ &AUC$\uparrow$ &ECE$\downarrow$ &BS$\downarrow$ &CECE$\downarrow$ &P-Value &95\%CI\\
        \midrule
        ShuffleNetV2~\cite{2018ShuffleNetV2} &78.18&54.71&55.99&88.59&4.51&29.27&22.40&\textless0.01&3.45&76.17&42.10&45.94&87.54&7.30&31.76&33.72&\textless0.01&6.01\\
        +ACB~\cite{2019ACNet} &80.91&55.68&57.88&88.76&5.05&25.27&24.09&\textless0.01&3.28&77.20&59.33&62.71&81.58&8.49&31.90&32.67&\textless0.01&5.92\\
        +DBB~\cite{2021DBB} &80.91&58.11&59.96&\textbf{88.92}&4.74&26.83&21.99&\textless0.01&3.28&78.76&56.00&55.41&89.26&7.96&28.81&34.87&\textless0.05&5.77\\
        +Rep~\cite{2021RepVGG} &80.18&56.72&58.65&88.16&\textbf{3.50}&26.97&21.28&\textless0.01&3.33&76.68&42.02&45.58&81.16&6.31&32.49&30.11&\textless0.01&5.97\\
        +MoRF~\cite{2025receptive} &78.18&51.64&51.45&87.38&3.70&29.69&26.94&\textless0.01&3.45&79.79&64.17&68.51&88.81&6.50&28.10&35.44&\textless0.05&5.67\\
        \textbf{+ERoHPRF} &\textbf{85.64}&\textbf{69.91}&\textbf{70.27}&88.25&6.17&\textbf{23.87}&\textbf{20.68}&-&\textbf{2.93}&\textbf{81.35}&\textbf{70.95}&\textbf{73.86}&\textbf{92.39}&\textbf{5.88}&\textbf{27.57}&\textbf{23.01}&-&\textbf{5.50}\\
        \hline
        MobileNetV2~\cite{2018MobileNetV2} &79.82&56.66&58.15&89.20&4.88&27.79&22.21&\textless0.01&3.35&76.68&40.30&41.99&81.72&5.02&31.57&30.31&\textless0.05&5.97\\
        +ACB~\cite{2019ACNet} &79.82&53.96&56.35&88.65&3.72&29.49&26.04&\textless0.01&3.37&79.28&59.53&60.63&83.87&7.87&29.25&31.00&\textless0.05&5.72\\
        +DBB~\cite{2021DBB} &80.55&54.23&55.73&88.46&5.21&28.60&26.80&\textless0.05&3.31&77.20&54.56&58.51&84.45&\textbf{4.74}&30.36&31.59&\textless0.05&5.92\\
        +Rep~\cite{2021RepVGG} &81.09&56.98&59.72&\textbf{89.20}&3.52&27.69&22.48&\textless0.05&3.27&76.68&40.88&44.00&83.41&6.57&31.92&30.16&\textless0.01&5.97\\
        +MoRF~\cite{2025receptive} &78.55&56.16&58.29&88.57&5.23&29.21&25.03&\textless0.01&3.43&78.24&51.48&51.28&84.39&7.14&30.01&34.53&\textless0.05&5.82\\
        \textbf{+ERoHPRF} &\textbf{84.00}&\textbf{63.41}&\textbf{66.13}&86.87&\textbf{3.01}&\textbf{24.03}&\textbf{20.22}&-&\textbf{3.06}&\textbf{84.46}&\textbf{68.39}&\textbf{71.65}&\textbf{85.31}&5.92&\textbf{26.17}&\textbf{28.53}&-&\textbf{5.11}\\
        \hline
        MixNet~\cite{2019MixConv} &79.64&57.09&59.47&86.39&5.36&30.03&28.37&\textless0.05&3.37&78.76&59.05&58.47&88.27&7.26&30.51&32.92&\textless0.05&5.77\\
        +ACB~\cite{2019ACNet} &81.27&59.05&62.00&87.98&4.72&27.49&27.54&\textless0.05&3.26&79.79&40.78&40.83&81.46&9.61&31.20&30.90&\textless0.01&5.67\\
        +DBB~\cite{2021DBB} &79.27&53.39&54.23&86.90&5.28&30.33&29.34&\textless0.01&3.39&77.72&55.06&56.97&85.69&6.18&32.48&33.21&\textless0.05&5.87\\
        +Rep~\cite{2021RepVGG} &79.64&54.49&57.32&87.86&5.81&27.84&26.22&\textless0.05&3.32&79.79&47.08&53.25&89.22&6.78&28.80&35.14&\textless0.05&5.67\\
        +MoRF~\cite{2025receptive} &80.36&56.82&59.35&88.44&7.16&28.14&25.62&\textless0.05&3.32&80.31&45.28&48.70&84.85&6.25&28.02&29.01&\textless0.01&5.61\\
        \textbf{+ERoHPRF} &\textbf{83.27}&\textbf{61.79}&\textbf{65.07}&\textbf{88.96}&\textbf{3.73}&\textbf{25.07}&\textbf{23.02}&-&\textbf{3.12}&\textbf{82.90}&\textbf{65.00}&\textbf{71.40}&\textbf{89.76}&\textbf{6.06}&\textbf{25.77}&\textbf{26.62}&-&\textbf{5.31}\\
        \bottomrule
    \end{tabular}}
    \label{table:sr}
\end{table*}

\subsection{Comparisons with SOTA Structural Reparameterization Methods}
Table~\ref{table:sr} presents the classification performance and fairness comparisons of our ERoHPRF and SOTA structural reparameterization methods across three efficient CNN backbones: ShuffleNetV2, MobileNeitV2, and MixNet. The model complexities of structural reparameterization methods based on MobileNetV2 are calculated on APTOS2019, as shown in \ref{table:efficiency}. Our ERoHPRF consistently keeps a better trade-off than competitive structural reparameterization methods in terms of medical image classification metrics and fairness metrics based on APTOS2019 (Left) and ISIC2018 (Right).

We first analyze the fair medical image classification results of our ERoHPRF and competitive structural reparameterization methods on the APTOS2019 dataset. For example, based on MobileNetV2, ERoHPRF outperforms DBB by \textbf{9.18\%} in bACC, \textbf{10.4\%} in mF1, and reduces 2.2\% in ECE, \textbf{4.57\%} in BS, and \textbf{6.58\%} in CECE, while slightly increasing the Params and MACs in the inference stage, as shown in Table~\ref{table:efficiency}. Remarkably, compared to MoRF, ERoHPRF achieves \textbf{18.27\%} gains of bACC, \textbf{18.82\%} gains of mF1, and obtains \textbf{5.82\%} and \textbf{6.26\%} reductions of BS and CECE based on ShuffleNetV2, manifesting that our ERoHPRF significantly improves the decision-making fairness and medical image classification performance of DNNs. Regarding the ISIC2018 dataset, our ERoHPRF outperforms ACB based on MixNet by \textbf{8.3\%} in AUC, as well as decreasing 3.55\% in ECE, \textbf{5.43\%} in BS, and \textbf{4.28\%} in CECE. Compared with Rep based on MobileNetV2, ERoHPRF obtains absolute over \textbf{27.51\%} and 1.9\% gains in bACC and AUC while decreasing over 0.65\%, \textbf{5.75\%}, and 1.63\% in ECE, BS, and CECE.

Additionally, the statistical significance tests prove the performance difference between ERoHPRF and other competitive methods. The results demonstrate that our method fully leverages the merits of heterogeneous pyramid RFs for improving the imbalanced classification performance and fairness from the multi-expert consultation perspective, agreeing with our expectations.

\begin{table*}[!t]
\caption{Performance comparisons between our ERoHPRF and SOTA DNNs in terms of medical image classification metrics, fairness metrics, and statistics metrics in the inference stage based on the APTOS2019 dataset and the ISIC2018 dataset.}
    \centering
    \resizebox{1.0\linewidth}{!}{
    \begin{tabular}{c|ccc|cccc|cc|ccc|cccc|cc} 
        \toprule
         \multirow{3}{*}{Method}&\multicolumn{9}{c|}{APTOS2019}&\multicolumn{9}{c}{ISIC2018}\\
         &\multicolumn{3}{c|}{Classification Metrics}&\multicolumn{4}{c|}{Fairness Metrics}&\multicolumn{2}{c|}{Statistical Significance}&\multicolumn{3}{c|}{Classification Metrics}&\multicolumn{4}{c}{Fairness Metrics}&\multicolumn{2}{c}{Statistical Significance}\\
         &ACC$\uparrow$ &bACC$\uparrow$ &mF1$\uparrow$ &AUC$\uparrow$ &ECE$\downarrow$ &BS$\downarrow$ &CECE$\downarrow$ &P-Value &95\%CI &ACC$\uparrow$ &bACC$\uparrow$ &mF1$\uparrow$ &AUC$\uparrow$ &ECE$\downarrow$ &BS$\downarrow$ &CECE$\downarrow$ &P-Value &95\%CI\\
        \midrule
        ResNet18~\cite{2016ResNet} &79.64&54.90&56.89&88.01&3.77&28.97&23.23&\textless0.01&3.37&77.72&58.01&56.88&83.29&7.27&29.71&29.97&\textless0.05&5.87\\
        VGG13~\cite{2015VGG} &80.36&54.32&57.06&88.75&3.63&28.38&25.28&\textless0.05&3.32&80.83&60.61&58.72&86.38&\textbf{4.07}&28.38&30.33&\textless0.05&5.55\\
        ConvNeXt~\cite{liu2022convnet} &71.46&35.91&31.01&82.72&4.89&39.73&41.69&\textless0.01&3.77&70.47&27.00&28.21&79.36&8.37&39.40&34.78&\textless0.01&6.44\\
        RepLKNet~\cite{2022RepLKNet} &80.73&54.43&56.67&\textbf{91.95}&4.54&26.58&27.33&\textless0.05&3.30&80.31&65.68&68.39&\textbf{92.53}&5.17&27.73&29.56&\textless0.05&5.61\\
        SLaKNet~\cite{2022SLaKNet} &75.64&39.42&36.06&83.12&5.91&36.34&39.28&\textless0.01&3.59&72.02&29.83&89.18&82.77&11.40&37.32&38.67&\textless0.01&6.33\\
        EdgeViT~\cite{pan2022edgevits} &82.00&59.14&61.98&91.10&4.46&26.26&23.50&\textless0.05&3.21&79.79&44.80&45.82&91.01&8.73&30.55&31.09&\textless0.05&5.67\\
        FasterNet~\cite{chen2023run} &77.82&52.12&53.93&85.63&4.95&31.50&25.42&\textless0.01&3.47&76.17&42.43&47.46&88.06&8.25&31.89&29.93&\textless0.05&6.01\\
        ViT~\cite{dosovitskiy2021image} &75.82&41.15&38.95&86.01&5.34&33.48&39.02&\textless0.01&3.58&73.06&37.44&40.06&82.94&7.75&36.74&31.00&\textless0.05&6.26\\
        Swin-T~\cite{liu2021swin} &75.46&45.90&47.01&87.19&3.28&33.46&28.41&\textless0.01&3.60&75.13&41.37&43.55&89.19&7.15&33.85&37.34&\textless0.05&6.10\\
        MLPMixer~\cite{tolstikhin2021mlp} &74.00&43.00&42.15&85.97&4.16&34.89&31.89&\textless0.01&3.67&80.83&52.45&55.50&91.76&9.90&28.53&34.15&\textless0.05&5.55\\
        Res-MLP~\cite{touvron2022resmlp}  &73.82&37.36&33.06&84.53&5.94&35.29&38.60&\textless0.01&3.67&74.09&46.76&47.60&81.74&7.41&34.81&37.44&\textless0.01&6.18\\
        \hline
        ShuffleNetV2~\cite{2018ShuffleNetV2}   &78.18&54.71&55.99&88.59&4.51&29.27&22.40&\textless0.01&3.45&76.17&42.10&45.94&87.54&7.30&31.76&33.72&\textless0.05&6.01\\
        MobileNetV2~\cite{2018MobileNetV2}    &79.82&56.66&58.15&89.20&4.88&27.79&22.21&\textless0.01&3.35&76.68&40.30&41.99&81.72&5.02&31.57&30.31&\textless0.05&5.97\\
        MixNet~\cite{2019MixConv}  &79.64&57.09&59.47&86.39&5.36&30.03&21.50&\textless0.01&3.37&78.76&59.05&58.47&88.27&7.26&30.51&32.92&\textless0.05&5.77\\
        EfficientNet~\cite{2019EfficientNet}  &81.09&55.11&57.93&89.37&5.26&28.40&28.37&\textless0.05&3.27&79.28&59.19&58.62&87.07&5.09&28.50&30.49&\textless0.05&5.72\\
        PVTv2~\cite{wang2022pvt}  &74.00&44.28&44.66&85.89&7.39&37.14&28.96&\textless0.01&3.67&73.06&33.19&34.57&83.14&5.03&36.12&32.79&\textless0.01&6.26\\
        StarNet~\cite{ma2024rewrite} &78.00&48.78&50.55&85.59&7.70&31.87&31.60&\textless0.01&3.46&77.20&43.20&46.74&89.30&4.40&31.41&35.76&\textless0.05&5.92\\
        \hline
        \textbf{ERoHPRF} &\textbf{84.00}&\textbf{63.41}&\textbf{66.13}&86.87&\textbf{3.01}&\textbf{24.03}&\textbf{20.22}&-&\textbf{3.06}&\textbf{84.46}&\textbf{68.39}&\textbf{71.65}&85.31&5.92&\textbf{26.17}&\textbf{28.53}&-&\textbf{5.11}\\
        \bottomrule
    \end{tabular}}
    \label{table:imbalance}
\end{table*}

\begin{table*}[!t]
\caption{Performance comparisons between our ERoHPRF and SOTA DNNs in terms of medical image classification metrics, fairness metrics, and statistics metrics in the inference stage based on the BTM dataset and the AS-OCT-NC dataset.}
    \centering
    \resizebox{1.0\linewidth}{!}{
    \begin{tabular}{c|ccc|cccc|cc|ccc|cccc|cc} 
        \toprule
         \multirow{3}{*}{Method}&\multicolumn{9}{c|}{BTM}&\multicolumn{9}{c}{AS-OCT-NC}\\
         &\multicolumn{3}{c|}{Classification Metrics}&\multicolumn{4}{c|}{Fairness Metrics}&\multicolumn{2}{c|}{Statistical Significance}&\multicolumn{3}{c|}{Classification Metrics}&\multicolumn{4}{c}{Fairness Metrics}&\multicolumn{2}{c}{Statistical Significance}\\
         &ACC$\uparrow$ &bACC$\uparrow$ &mF1$\uparrow$ &AUC$\uparrow$ &ECE$\downarrow$ &BS$\downarrow$ &CECE$\downarrow$ &P-Value &95\%CI &ACC$\uparrow$ &bACC$\uparrow$ &mF1$\uparrow$ &AUC$\uparrow$ &ECE$\downarrow$ &BS$\downarrow$ &CECE$\downarrow$ &P-Value &95\%CI\\
        \midrule
        ResNet18~\cite{2016ResNet} &75.64&73.21&73.18&89.37&15.85&38.43&19.86&\textless0.01&4.24&89.08&74.52&79.65&94.96&13.36&21.39&19.16&\textless0.01&0.99\\
        VGG13~\cite{2015VGG} &77.92&76.55&76.68&89.62&12.72&35.61&16.03&\textless0.01&4.10&88.24&75.29&80.82&95.03&15.32&19.08&23.01&\textless0.01&1.03\\
        ConvNeXt~\cite{liu2022convnet} &79.70&79.12&79.01&91.22&12.22&34.80&16.68&\textless0.05&3.97&89.35&77.02&82.18&92.52&15.96&23.47&17.75&\textless0.01&0.98\\
        RepLKNet~\cite{2022RepLKNet} &80.20&78.97&79.12&91.68&14.67&33.53&17.61&\textless0.05&3.93&89.14&79.72&83.92&95.22&11.57&20.02&20.13&\textless0.01&0.99\\
        SLaKNet~\cite{2022SLaKNet} &75.13&74.62&74.76&90.31&11.86&37.13&18.12&\textless0.01&4.27&88.87&75.93&81.05&76.74&12.90&22.59&18.19&\textless0.01&1.00\\
        EdgeViT~\cite{pan2022edgevits} &77.92&76.93&77.18&88.52&13.27&37.34&17.59&\textless0.01&4.10&87.98&69.05&71.58&95.22&13.40&21.56&23.85&\textless0.01&1.04\\
        FasterNet~\cite{chen2023run} &73.60&71.35&71.20&81.49&16.97&43.16&21.79&\textless0.01&4.35&87.50&75.30&80.99&94.73&12.28&23.47&17.80&\textless0.01&1.05\\
        ViT~\cite{dosovitskiy2021image} &75.38&73.23&73.86&89.65&14.13&38.29&17.10&\textless0.01&4.25&80.59&69.62&73.77&91.04&14.20&33.05&22.94&\textless0.01&1.26\\
        Swin-T~\cite{liu2021swin} &78.68&77.14&77.48&91.12&14.37&33.67&17.68&\textless0.01&4.04&90.03&79.09&83.41&95.11&17.27&23.05&17.35&\textless0.01&0.95\\
        MLPMixer~\cite{tolstikhin2021mlp} &76.40&74.39&74.14&89.61&16.11&38.00&20.27&\textless0.01&4.19&89.93&77.81&82.84&93.49&17.80&22.47&19.57&\textless0.01&0.96\\
        Res-MLP~\cite{touvron2022resmlp} &75.89&74.60&74.19&89.28&13.99&37.89&18.05&\textless0.01&4.22&86.60&71.14&76.62&66.96&16.91&26.12&17.42&\textless0.01&1.08\\
        \hline
        ShuffleNetV2~\cite{2018ShuffleNetV2} &80.46&79.18&79.46&90.61&16.73&35.57&19.14&\textless0.05&3.92&89.72&79.76&84.42&95.31&15.72&22.99&22.03&\textless0.01&0.97\\
        MobileNetV2~\cite{2018MobileNetV2} &80.20&78.50&78.33&91.01&12.44&34.91&17.49&\textless0.05&3.93&88.58&78.10&82.06&94.70&14.98&22.90&18.61&\textless0.01&1.01\\
        MixNet~\cite{2019MixConv} &81.47&80.49&80.29&89.58&15.33&34.05&19.00&\textless0.05&3.84&88.82&77.22&82.48&95.25&11.58&21.90&18.12&\textless0.01&1.00\\
        EfficientNet~\cite{2019EfficientNet} &78.68&76.64&76.30&90.94&14.78&37.34&20.82&\textless0.01&4.04&89.27&80.25&83.91&94.92&12.95&20.15&17.50&\textless0.01&0.99\\
        PVTv2~\cite{wang2022pvt} &71.07&68.62&69.40&87.35&13.52&43.65&19.81&\textless0.01&4.48&89.11&78.98&82.97&94.77&19.28&24.87&21.67&\textless0.01&0.99\\
        StarNet~\cite{ma2024rewrite} &80.20&78.83&79.11&91.44&13.04&34.59&17.38&\textless0.05&3.93&90.06&78.27&82.27&92.47&23.33&25.14&24.13&\textless0.01&0.95\\
        \hline
        \textbf{ERoHPRF} &\textbf{83.76}&\textbf{83.36}&\textbf{83.33}&\textbf{91.83}&\textbf{11.66}&\textbf{27.96}&\textbf{14.04}&-&\textbf{3.64}&\textbf{91.69}&\textbf{83.04}&\textbf{86.01}&\textbf{95.39}&\textbf{10.86}&\textbf{17.72}&\textbf{17.08}&-&\textbf{0.88}\\
        \bottomrule
    \end{tabular}}
    \label{table:balance}
\end{table*}

\subsection{Comparisons with SOTA Deep Neural Networks}
This paper further investigates the effectiveness, fairness, and efficiency of ERoHPRF through comparisons to SOTA DNNs based on four imbalanced/balanced datasets. We choose MobileNetV2 as the backbone, because our ERoHPRF achieves the best overall performance in Table~\ref{table:sr} based on MobileNetV2. The model complexities of our ERoHPRF and SOTA DNNs are presented in Table~\ref{table:efficiency} based on APTOS2019, including the training and inference stages.

\textbf{Results on APTOS2019 and ISIC2018.} Table~\ref{table:imbalance} presents the imbalanced medical image classification, fairness, and statistics comparisons of our ERoHPRF and other advanced DNNs on APTOS2019 (Left) and ISIC2018 (Right). Table~\ref{table:efficiency} presents the model complexity of our ERoHPRF and SOTA DNNs based on APTOS2019. It observes that ERoHPRF obtains a better trade-off in terms of imbalanced classification results, fairness, and model complexity. We first compare our ERoHPRF to advanced DNNs on the APTOS2019 dataset, as shown in Table \ref{table:imbalance} (Left). Compared with RepLKNet, ERoHPRF achieves over \textbf{8.98\%} and \textbf{9.46\%} gains of bACC and mF1 with a reduction of 1.53\%, 2.55\%, and \textbf{7.11\%} in ECE, BS, and CECE, while reducing absolute over \textbf{96.8\%} and \textbf{97.36\%} in Params and MACs in the inference stage. Remarkably, ERoHPRF significantly outperforms SLaKNet by \textbf{8.36\%} in ACC and 3.75\% in AUC, while reducing BS and CECE by \textbf{12.31\%} and \textbf{19.06\%} with \textbf{93.33\%} and \textbf{95.6\%} reductions in Params and MACs. Similarly, based on ISIC2018, our ERoHPRF generally performs better than other SOTA DNNs. Compared with ViT and Swin-T, our ERoHPRF achieves absolute over \textbf{9.33\%}, \textbf{27.02\%}, and \textbf{28.1\%} gains in ACC, bACC, and mF1, while reducing ECE, BS, and CECE by 1.23\%, \textbf{7.68\%}, and 2.47\%. ERoHRPF also outperforms StarNet by \textbf{25.19\%}, \textbf{24.91\%}, \textbf{5.24\%}, and \textbf{7.23\%} in bACC, mF1, BS, and CECE. Overall, the results prove the effectiveness, fairness, and efficiency of mining the potential of the heterogeneous pyramid RFs to improve imbalanced medical image classification performance and fairness of DNNs, aligning with our motivation.

\textbf{Results on BTM and AS-OCT-NC.} Table~\ref{table:balance} offers balanced medical image classification and fairness comparisons of our ERoHPRF and other advanced DNNs on BTM (Left) and AS-OCT-NC (Right). ERoHPRF generally performs better than other competitive DNNs in terms of medical image classification metrics and fairness metrics. We conclude: (1) Result comparisons based on the BTM dataset. Compared with EdgeViT, ERoHPRF obtains \textbf{6.43\%}, \textbf{6.15\%}, and 3.31\% gains in bACC, mF1 and AUC, while achieving 1.61\%, \textbf{9.38\%}, and 3.55\% reductions in ECE, BS, and CECE. ERoHPRF also outperforms PVTv2 and StarNet by over 3.56\%, \textbf{4.53\%}, and 0.39\% in ACC, bACC, and AUC, while decreasing the ECE, BS, and CECE by over 1.38\%, \textbf{6.63\%}, and 3.34\% accordingly. (2) Result comparisons on the AS-OCT-NC dataset. ERoHPRF obtains above absolute 2.55\%, 3.32\%, and 2.09\% gains in ACC, bACC, and mF1, while achieving 0.71\%, 2.3\%, and 1.11\% reductions in ECE, BS, and CECE, through comparisons to RepLKNet and SLaKNet. Besides, ERoHPRF outperforms EdgeViT and FasterNet by over absolute 3.71\%, \textbf{7.74\%}, and \textbf{5.02\%} in ACC, bACC, and mF1, as well as obtaining 1.42\%, 3.84\%, and 0.72\% reductions of ECE, BS, and CECE, respectively.

Moreover, statistical significance tests in Table~\ref{table:imbalance} and Table~\ref{table:balance} verify the learned feature representation difference among our ERoHPRF and other SOTA DNNs, demonstrating our motivation to improve imbalanced medical image classification performance and fairness from the perspective of heterogeneous pyramid RFs.

\subsection{Efficiency and Fairness Analysis}

\textbf{Efficiency Analysis and Visualization.} Table~\ref{table:efficiency} offers the model complexity and speed in both training and inference stages based on APTOS2019. All methods in the training stage are conducted on a server with an NVIDIA RTX 3090 GPU and an AMD EPYC 7402 CPU, as introduced in Section~\ref{exp-setting}. \textbf{We also adopt a laptop with an AMD Ryzen 7840HS CPU as a resource-constrained medical device to test the model complexity and speed of all methods in the inference stage.} We observe as follows: in the training stage, compared to advanced CNNs, Transformers, and MLP-like architectures, ERoHPRF requires smaller VRAM with fewer parameters and MACs. As for the inference stage, ERoHPRF shows a competitive speed with a better trade-off in classification and fairness performance and computation overhead. Compared to RepLKNet, our ERoHPRF is \textbf{27} times faster while reducing \textbf{96.8\%} and \textbf{97.36\%} in Params and MACs. Our ERoHPRF is also \textbf{2.43} times and \textbf{1.26} times faster than EdgeViT and FasterNet, meeting the requirement of resource-constrained medical devices.

Table~\ref{table:infertime} further offers computational overhead comparison of our ERoHPRF and other representative structure reparameterization methods based on inference time: ACB, DBB, Rep, MoRF, RepLKNet, and SLaKNet. The inference time is calculated on a laptop with an AMD Ryzen 7840HS CPU based on the APTOS2019 dataset. We observe that our ERoHPRF keeps the competitive inference time through comparisons to other structural reparameterization methods, also meeting the requirements of the real-time clinical diagnosis, and can be deployed on resource-limited medical devices.

\begin{table}
\caption{Efficiency comparisons between our ERoHPRF and SOTA DNNs in terms of training complexity, training speed, inference complexity, and inference speed based on the APTOS2019 dataset.}
    \centering
    \resizebox{1.0\linewidth}{!}{
    \begin{tabular}{c|ccc|ccc} 
        \toprule
         \multirow{2}{*}{Method}&\multicolumn{3}{c}{Training (Server)}&\multicolumn{3}{c}{Inference (Edge)}\\
         &Params$\downarrow$ &MACs$\downarrow$ &FPS$\uparrow$ &Params$\downarrow$ &MACs$\downarrow$ &FPS$\uparrow$\\
        \midrule
        ResNet18~\cite{2016ResNet} &11.179M&2.382G&396&11.179M&2.382G&29.2\\
        VGG13~\cite{2015VGG} &9.413M&14.673G&\textbf{494}&9.413M&14.673G&8.2\\
        ConvNeXt~\cite{liu2022convnet} &27.803M&5.818G&194&27.803M&5.818G&7.4\\
        RepLKNet~\cite{2022RepLKNet} &78.844M&20.369G&45&78.507M&20.235G&1.0\\
        SLaKNet~\cite{2022SLaKNet} &37.819M&12.237G&103&37.634M&12.145G&0.7\\
        EdgeViT~\cite{pan2022edgevits} &12.725M&2.478G&83&12.725M&2.478G&11.1\\
        FasterNet~\cite{chen2023run} &13.708M&2.502G&218&13.708M&2.502G&21.4\\
        ViT~\cite{dosovitskiy2021image} &53.543M&3.479G&258&53.543M&3.479G&14.1\\
        Swin-T~\cite{liu2021swin} &27.500M&5.712G&99&27.500M&5.712G&6.5\\
        MLPMixer~\cite{tolstikhin2021mlp} &18.264M&4.941G&296&18.264M&4.941G&10.3\\
        Res-MLP~\cite{touvron2022resmlp} &15.266M&4.001G&249&15.266M&4.001G&11.9\\
        \hline
        ShuffleNetV2~\cite{2018ShuffleNetV2} &5.328M&0.753G&137&5.328M&0.753G&\textbf{35.8}\\
        MixNet~\cite{2019MixConv} &2.605M&\textbf{0.343G}&93&2.605M&\textbf{0.343G}&19.0\\
        EfficientNet~\cite{2019EfficientNet} &2.910M&1.844G&127&2.910M&1.844G&8.3\\
        PVTv2~\cite{wang2022pvt} &3.411M&0.695G&155&3.411M&0.695G&20.8\\
        StarNet~\cite{ma2024rewrite} &2.677M&0.558G&158&2.677M&0.558G&35.2\\
        \hline
        MobileNetV2~\cite{2018MobileNetV2} &\textbf{2.230M}&0.426G&161&2.230M&0.426G&26.6\\
        +ACB~\cite{2019ACNet} &2.302M&0.468G&111&2.223M&0.414G&26.7\\
        +DBB~\cite{2021DBB} &2.430M&0.556G&67&2.223M&0.414G&26.8\\
        +Rep~\cite{2021RepVGG} &2.264M&0.451G&113&\textbf{2.223M}&0.414G&26.4\\
        +MoRF~\cite{2025receptive} &2.243M&0.604G&137&2.240M&0.586G&26.4\\
        \textbf{+ERoHPRF} &3.929M&1.226G&37&2.509M&0.534G&27.0\\
        \hline
    \end{tabular}}
    \label{table:efficiency}
\end{table}

\begin{table}
\caption{Computational overhead comparisons of ERoHPRF and other representative structure reparameterization methods based on the inference time.}
    \begin{center}
    \resizebox{1.0\linewidth}{!}{
    \begin{tabular}{c|c|c|c|c|c|c|c}
    \hline
    Method&ACB&DBB&Rep&MoRF&RepLKNet&SLaKNet&ERoHPRF\\
    \hline 
    Time (ms)&37.5&37.3&37.9&37.9&1000&1429&\textbf{37.0}\\
    \hline
    \end{tabular}}
    \end{center}
    \label{table:infertime}
\end{table}

\begin{figure}[t]
\centering
\includegraphics[width=1.0\linewidth]{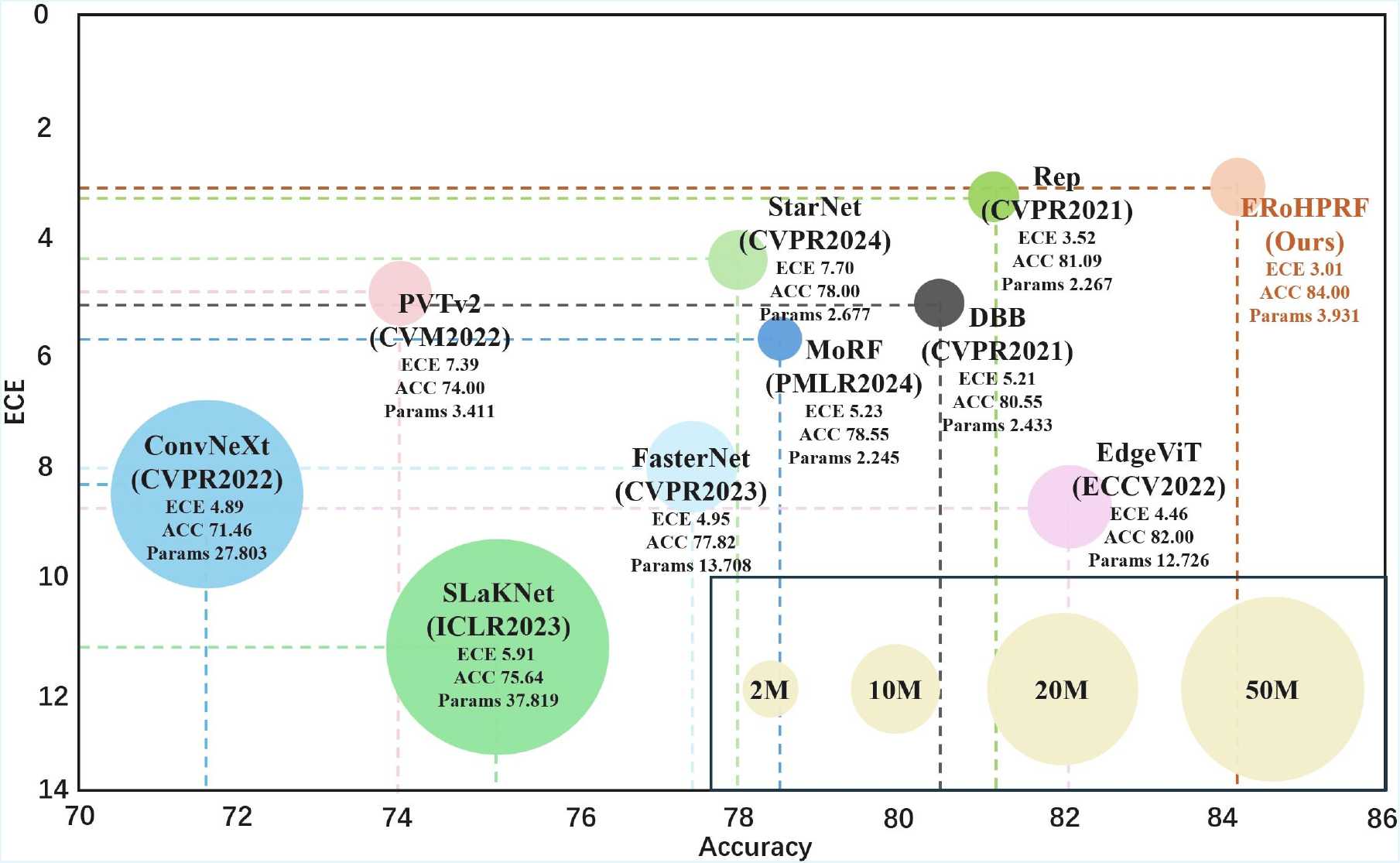}
 \caption{Comprehensive Comparisons of our ERoHPRF, existing state-of-the-art efficient DNNs, and structural reparameterization methods in terms of accuracy ($\uparrow$), expected calibration error (ECE) ($\downarrow$), and parameter number ($\downarrow$) based on the APTOS2019 dataset. The bubble size corresponds to the parameter number of all methods in the training stage. The larger the bubble size, the greater the method's parameter number.}
   \label{TrainingCost}
\end{figure}

Fig.~\ref{InferenceCost} and Fig.~\ref{TrainingCost} provide the efficiency comparisons of our ERoHPRF and other advanced methods in the inference and training stages based on the APTOS2019 dataset. It can be observed that our ERoHPRF achieves a better trade-off in terms of medical image classification, fairness, and complexity both in the training and inference stages. In the training stage, ERoHPRF obtains the best classification and fairness results without significantly increasing the number of parameters. Regarding the inference stage, ERoHPRF employs the expert-like structural reparameterization method to merge the parameters of heterogeneous pyramid receptive fields into an extensive RF; thus, our method further reduces computational overhead without degrading the performance of medical image classification and fairness, showing the superiority of ERoHPRF over other SOTA methods, which has excellent potential to be deployed on resource-constrained medical devices.

\begin{table*}
\caption{Performance comparisons of our ERoHPRF and other advanced DNNs from the age subgroup aspect based on the ISIC2018 dataset and the AS-OCT-NC dataset.}
    \centering
    \resizebox{1.0\linewidth}{!}{
    \begin{tabular}{c|c|ccc|cccc|ccc|cccc} 
        \toprule
         \multirow{3}{*}{Subgroup}&\multirow{3}{*}{Method}&\multicolumn{7}{c|}{ISIC2018}&\multicolumn{7}{c}{AS-OCT-NC}\\
         &&\multicolumn{3}{c|}{Classification Metrics}&\multicolumn{4}{c|}{Fairness Metrics}&\multicolumn{3}{c|}{Classification Metrics}&\multicolumn{4}{c}{Fairness Metrics}\\
         &&ACC$\uparrow$ &bACC$\uparrow$ &mF1$\uparrow$ &AUC$\uparrow$ &ECE$\downarrow$ &BS$\downarrow$ &CECE$\downarrow$&ACC$\uparrow$ &bACC$\uparrow$ &mF1$\uparrow$ &AUC$\uparrow$ &ECE$\downarrow$ &BS$\downarrow$ &CECE$\downarrow$\\
        \midrule
        \multirow{5}{*}{\makecell{Age in \\ $[0, 50]$}}
        &VGG13~\cite{2015VGG} &98.25&99.11&82.88&66.96&7.03&6.70&38.42&81.55&59.87&62.88&100.00&15.51&31.35&8.49\\
        &RepLKNet~\cite{2022RepLKNet} &98.25&99.11&82.88&85.43&7.08&6.25&27.46&91.83&62.13&64.88&100.00&16.69&15.47&9.31\\
        &EdgeViT~\cite{pan2022edgevits} &96.49&32.74&33.03&70.54&9.36&8.67&26.64&44.05&46.05&51.75&100.00&15.77&52.64&8.86\\
        &EfficientNet~\cite{2019EfficientNet} &96.49&32.74&33.03&80.36&7.13&5.91&27.64&89.29&62.72&64.57&100.00&16.74&21.91&9.72\\
        &ERoHPRF &\textbf{98.25}&\textbf{99.11}&\textbf{82.88}&\textbf{85.71}&\textbf{6.83}&\textbf{5.84}&\textbf{26.38}&\textbf{92.26}&\textbf{63.82}&\textbf{65.18}&\textbf{100.00}&\textbf{15.18}&\textbf{14.48}&\textbf{8.39}\\
        \hline
        \multirow{5}{*}{\makecell{Age in \\ $[51, 69]$}}
        &VGG13~\cite{2015VGG} &78.16&65.57&63.35&86.76&11.79&31.66&38.00&82.43&56.98&54.04&81.83&20.08&31.20&35.83\\
        &RepLKNet~\cite{2022RepLKNet} &75.86&66.06&66.36&85.89&\textbf{11.52}&34.81&33.44&82.58&56.92&54.50&82.18&20.94&32.20&36.52\\
        &EdgeViT~\cite{pan2022edgevits} &77.01&45.79&44.84&86.33&14.87&34.17&34.21&89.16&62.14&60.72&82.08&23.43&28.73&36.46\\
        &EfficientNet~\cite{2019EfficientNet} &77.01&65.06&61.06&85.84&11.85&33.26&36.72&86.61&60.69&58.99&82.40&20.18&29.06&35.43\\
        &ERoHPRF &\textbf{83.91}&\textbf{75.87}&\textbf{76.81}&\textbf{86.93}&12.25&\textbf{29.35}&\textbf{33.39}&\textbf{90.80}&\textbf{62.82}&\textbf{62.15}&\textbf{82.70}&\textbf{19.01}&\textbf{24.42}&\textbf{35.32}\\
        \hline
        \multirow{5}{*}{\makecell{Age in \\ $[70, 100]$}}
        &VGG13~\cite{2015VGG} &64.58&49.50&45.46&83.25&15.05&48.08&37.79&92.57&84.35&86.09&94.85&8.87&12.87&23.33\\
        &RepLKNet~\cite{2022RepLKNet} &66.67&55.65&54.77&\textbf{84.72}&16.25&45.39&37.21&92.29&84.67&85.43&95.37&11.08&12.31&25.44\\
        &EdgeViT~\cite{pan2022edgevits} &64.58&49.50&44.78&83.93&15.48&50.55&40.56&92.94&84.71&85.72&94.79&11.48&12.81&16.95\\
        &EfficientNet~\cite{2019EfficientNet} &62.50&48.06&47.59&83.03&15.14&47.40&35.75&91.81&85.08&84.28&95.25&8.47&13.35&17.11\\
        &\textbf{ERoHPRF} &\textbf{68.75}&\textbf{56.16}&\textbf{55.19}&82.73&\textbf{14.97}&\textbf{45.08}&\textbf{31.25}&\textbf{93.27}&\textbf{85.29}&\textbf{86.37}&\textbf{95.54}&\textbf{6.62}&\textbf{12.10}&\textbf{16.14}\\
        \bottomrule
    \end{tabular}}
    \label{table:fairness_age}
\end{table*}

\begin{table*}
\caption{Performance comparisons of our ERoHPRF and other advanced DNNs from the sex subgroup aspect based on the ISIC2018 dataset and the AS-OCT-NC dataset.}
    \centering
    \resizebox{1.0\linewidth}{!}{
    \begin{tabular}{c|c|ccc|cccc|ccc|cccc} 
        \toprule
         \multirow{3}{*}{Subgroup}&\multirow{3}{*}{Method}&\multicolumn{7}{c|}{ISIC2018}&\multicolumn{7}{c}{AS-OCT-NC}\\
         &&\multicolumn{3}{c|}{Classification Metrics}&\multicolumn{4}{c|}{Fairness Metrics}&\multicolumn{3}{c|}{Classification Metrics}&\multicolumn{4}{c}{Fairness Metrics}\\
         &&ACC$\uparrow$ &bACC$\uparrow$ &mF1$\uparrow$ &AUC$\uparrow$ &ECE$\downarrow$ &BS$\downarrow$ &CECE$\downarrow$&ACC$\uparrow$ &bACC$\uparrow$ &mF1$\uparrow$ &AUC$\uparrow$ &ECE$\downarrow$ &BS$\downarrow$ &CECE$\downarrow$\\
        \midrule
        \multirow{5}{*}{Male}
        &VGG13~\cite{2015VGG} &83.62&56.41&59.52&82.23&8.77&28.95&36.67&88.89&72.98&79.93&94.45&12.63&17.85&21.33\\
        &RepLKNet~\cite{2022RepLKNet} &82.76&63.34&62.64&82.94&8.30&28.62&32.87&88.53&75.66&82.00&94.72&12.03&19.06&25.75\\
        &EdgeViT~\cite{pan2022edgevits} &79.31&47.05&47.21&83.06&9.26&31.33&33.46&88.07&66.83&72.45&94.66&10.21&18.36&21.29\\
        &EfficientNet~\cite{2019EfficientNet} &78.45&51.16&51.89&81.86&\textbf{8.04}&31.55&32.18&88.80&78.21&83.14&94.12&10.18&17.97&21.44\\
        &ERoHPRF &\textbf{83.62}&\textbf{65.35}&\textbf{63.25}&\textbf{83.21}&8.47&\textbf{27.95}&\textbf{31.48}&\textbf{90.62}&\textbf{78.59}&\textbf{83.94}&\textbf{94.75}&\textbf{10.16}&\textbf{17.81}&\textbf{20.80}\\
        \hline
        \multirow{5}{*}{Female}
        &VGG13~\cite{2015VGG} &76.32&41.61&35.84&83.02&9.40&27.44&39.06&89.02&78.01&83.28&94.75&13.34&19.94&17.61\\
        &RepLKNet~\cite{2022RepLKNet} &76.32&45.48&42.98&\textbf{84.29}&10.14&28.29&39.94&89.96&82.16&85.43&94.78&13.91&19.77&17.95\\
        &EdgeViT~\cite{pan2022edgevits} &80.26&36.61&33.00&83.58&10.20&29.72&39.67&89.81&71.47&74.50&94.97&16.23&21.45&22.34\\
        &EfficientNet~\cite{2019EfficientNet} &80.26&49.41&45.16&83.08&10.74&24.22&40.03&90.20&82.07&85.50&95.13&14.99&20.21&17.76\\
        &\textbf{ERoHPRF} &\textbf{85.53}&\textbf{52.20}&\textbf{55.43}&82.84&\textbf{8.24}&\textbf{23.79}&\textbf{38.83}&\textbf{93.05}&\textbf{85.66}&\textbf{88.04}&\textbf{95.84}&\textbf{13.16}&\textbf{16.29}&\textbf{17.20}\\
        \bottomrule
    \end{tabular}}
    \label{table:fairness_gender}
\end{table*}

\begin{table*}
\caption{Performance comparisons of our ERoHPRF and other advanced DNNs from the disease subgroup aspect based on the ISIC2018 dataset and the APTOS2019 dataset.}
    \centering
    \resizebox{1.0\linewidth}{!}{
    \begin{tabular}{c|c|ccc|cccc|ccc|cccc} 
        \toprule
         \multirow{3}{*}{Subgroup}&\multirow{3}{*}{Method}&\multicolumn{7}{c|}{ISIC2018}&\multicolumn{7}{c}{APTOS2019}\\
         &&\multicolumn{3}{c|}{Classification Metrics}&\multicolumn{4}{c|}{Fairness Metrics}&\multicolumn{3}{c|}{Classification Metrics}&\multicolumn{4}{c}{Fairness Metrics}\\
         &&ACC$\uparrow$ &bACC$\uparrow$ &mF1$\uparrow$ &AUC$\uparrow$ &ECE$\downarrow$ &BS$\downarrow$ &CECE$\downarrow$&ACC$\uparrow$ &bACC$\uparrow$ &mF1$\uparrow$ &AUC$\uparrow$ &ECE$\downarrow$ &BS$\downarrow$ &CECE$\downarrow$\\
        \midrule
        \multirow{5}{*}{Head}
        &VGG13~\cite{2015VGG} &86.15&42.35&43.34&78.87&\textbf{6.69}&21.17&20.74&93.56&36.76&37.74&98.48&9.60&13.09&14.16\\
        &RepLKNet~\cite{2022RepLKNet} &84.34&40.28&40.86&79.22&6.72&21.88&21.32&92.87&35.20&36.88&98.81&6.80&11.24&10.50\\
        &EdgeViT~\cite{pan2022edgevits} &86.75&55.06&55.82&\textbf{79.88}&11.22&23.42&22.41&93.33&36.76&37.74&98.43&6.57&12.12&12.52\\
        &EfficientNet~\cite{2019EfficientNet} &84.94&33.77&35.70&78.98&6.91&21.51&21.20&93.79&36.76&38.10&99.20&6.90&11.79&11.77\\
        &ERoHPRF &\textbf{87.95}&\textbf{55.52}&\textbf{58.50}&78.36&7.31&\textbf{20.67}&\textbf{20.65}&\textbf{93.79}&\textbf{36.76}&\textbf{38.10}&\textbf{99.35}&\textbf{6.40}&\textbf{11.18}&\textbf{10.26}\\
        \hline
        \multirow{5}{*}{Tail}
        &VGG13~\cite{2015VGG} &48.15&30.36&35.19&84.68&27.72&72.71&39.02&30.44&17.57&25.13&78.42&26.50&86.18&32.70\\
        &RepLKNet~\cite{2022RepLKNet} &55.56&42.06&47.97&84.17&30.58&63.76&35.07&34.78&18.27&25.12&84.01&34.91&88.39&41.88\\
        &EdgeViT~\cite{pan2022edgevits} &37.04&13.33&16.01&76.59&27.26&74.39&35.56&39.13&22.26&29.66&83.45&25.53&78.26&32.42\\
        &EfficientNet~\cite{2019EfficientNet} &44.44&30.24&35.71&80.94&27.19&71.48&37.46&33.04&18.16&25.84&78.42&31.15&91.22&39.43\\
        &\textbf{ERoHPRF} &\textbf{62.96}&\textbf{42.78}&\textbf{48.35}&\textbf{85.92}&\textbf{26.63}&\textbf{60.04}&\textbf{34.45}&\textbf{46.96}&\textbf{26.65}&\textbf{34.41}&\textbf{85.63}&\textbf{23.10}&\textbf{72.65}&\textbf{26.85}\\
        \bottomrule
    \end{tabular}}
    \label{table:fairness_disease}
\end{table*}

\textbf{Fairness Analysis and Visualization.} Table~\ref{table:fairness_age}, Table~\ref{table:fairness_gender}, and Table~\ref{table:fairness_disease} offer the fairness trade-off analysis among ERoHPRF and the other four representative DNNs from age, sex, and disease subgroup analysis aspects based on the imbalanced ISIC2018 dataset, the imbalanced APTOS2019 dataset, and the balanced AS-OCT-NC dataset. Due to the missing values in age and sex, we only select 192 dermatoscopic images from the ISIC2018 dataset and 3,629 AS-OCT images from the AS-OCT-NC dataset for Table~\ref{table:fairness_age} and Table~\ref{table:fairness_gender}. Table~\ref{table:fairness_disease} uses 193 dermatoscopic images in ISIC2018 and 550 fundus images in APTOS2019, consistent with  Table~\ref{table:imbalance} and Table~\ref{table:sr}.

Regarding the age subgroup aspect, we divide the age into three subgroups according to the age distribution: $[0,50]$ (ISIC2018: 57, AS-OCT-NC: 168), $[51,69]$ (ISIC2018: 87, AS-OCT-NC: 1,337), and $[70,100]$ (ISIC2018: 48, AS-OCT-NC: 2,124). Our ERoHPRF generally performs a better fairness trade-off than other DNNs across three age subgroups. For example, compared to EdgeViT, our ERoHPRF achieves \textbf{30.08\%} and \textbf{31.97\%} gains in bACC and mF1, while obtaining \textbf{4.82\%} and 0.82\% reductions in BS and CECE in subgroup $[51,69]$ on the ISIC2018 dataset. ERoHPRF also outperforms EfficientNet by 2.09\%, 1.85\%, 1.25\%, and 0.97\% in mF1, ECE, BS, and CECE in subgroup $[70,100]$ on the AS-OCT-NC dataset.

As for the sex subgroup aspect in Table~\ref{table:fairness_gender}, we divide the sex into two subgroups: Male (ISIC2018: 116, AS-OCT-NC: 1,098) and Female (ISIC2018: 76, AS-OCT-NC: 2,531). Compared to VGG13 in the Male subgroup on ISIC2018, our ERoHPRF obtains \textbf{8.94\%} and 3.73\% gains in bACC and mF1, while decreasing 1\% and \textbf{5.19\%} in BS and CECE. Our ERoHPRF also outperforms EdgeViT in ACC, bACC, BS, and CECE by 3.24\%, \textbf{14.19\%}, \textbf{5.16\%}, and \textbf{5.14\%} in the Female subgroup on AS-OCT-NC. Overall, ERoHPRF obtains a better trade-off than competitive DNNs in fairness metrics, aligning with our expectations.

Regarding the disease subgroup aspect, we select the classes with fewer than 20 images as the tail subgroup (4 classes, 27 images) and the rest as the head subgroup (3 classes, 166 images) based on the ISIC2018 by following the literature~\cite{holste2022long}. Similarly, we choose the classes with fewer than 100 images as the tail subgroup (3 classes, 115 images) and the rest as the head subgroup (2 classes, 435 images) on the APTOS2019. Compared to VGG13,  ERoHPRF achieves \textbf{14.81\%} and \textbf{13.16\%} improvements of ACC and mF1, while reducing \textbf{12.67\%} and \textbf{4.57\%} in BS and CECE in the tail subgroup on ISIC2018. Our ERoHPRF also outperforms RepLKNet by \textbf{8.38\%} in bACC, \textbf{9.29\%} in mF1, \textbf{11.81\%} in ECE, and \textbf{15.03\%} in CECE in the tail subgroup on APTOS2019. Generally, our ERoHPRF maintains a better fairness trade-off than other DNNs in classifying different disease groups, agreeing with our expectations.

\begin{figure*}[!t]
\centering
\includegraphics[width=0.95\linewidth]{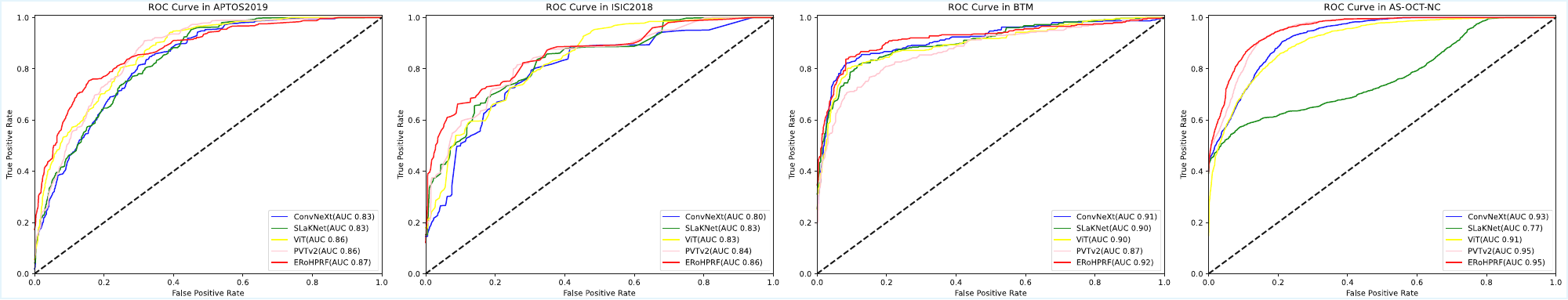}
 \caption{The ROC curves of our ERoHPRF, three advanced DNNs, and one structural reparameterization method based on the APTOS2019 dataset, the ISIC2018 dataset, the BTM dataset, and the AS-OCT-NC dataset.}
\label{AUC}
\end{figure*}

\begin{figure*}[!t]
\centering
\includegraphics[width=0.95\linewidth]{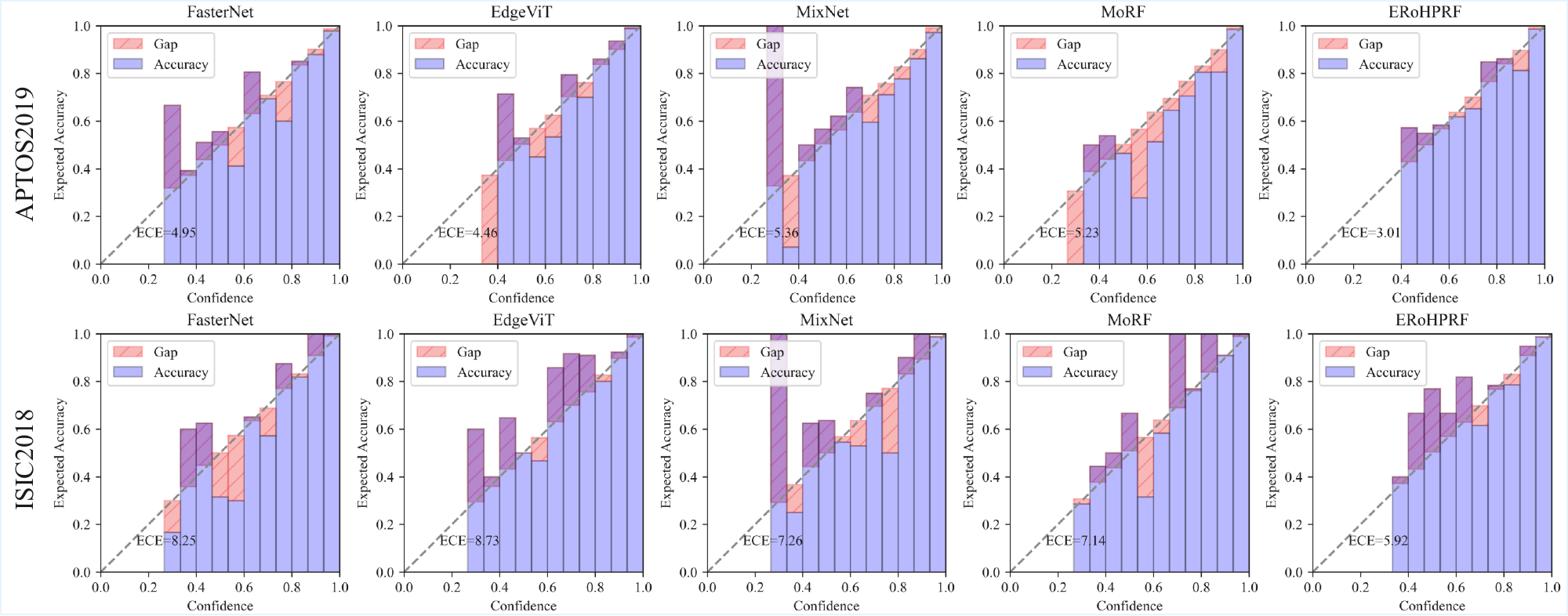}
 \caption{The reliability diagrams of our ERoHPRF, three advanced DNNs, and one structural reparameterization method on the APTOS2019 dataset and the ISIC2018 dataset.}
\label{ECE_ai}
\end{figure*}

\begin{figure*}
\centering
\includegraphics[width=0.95\linewidth]{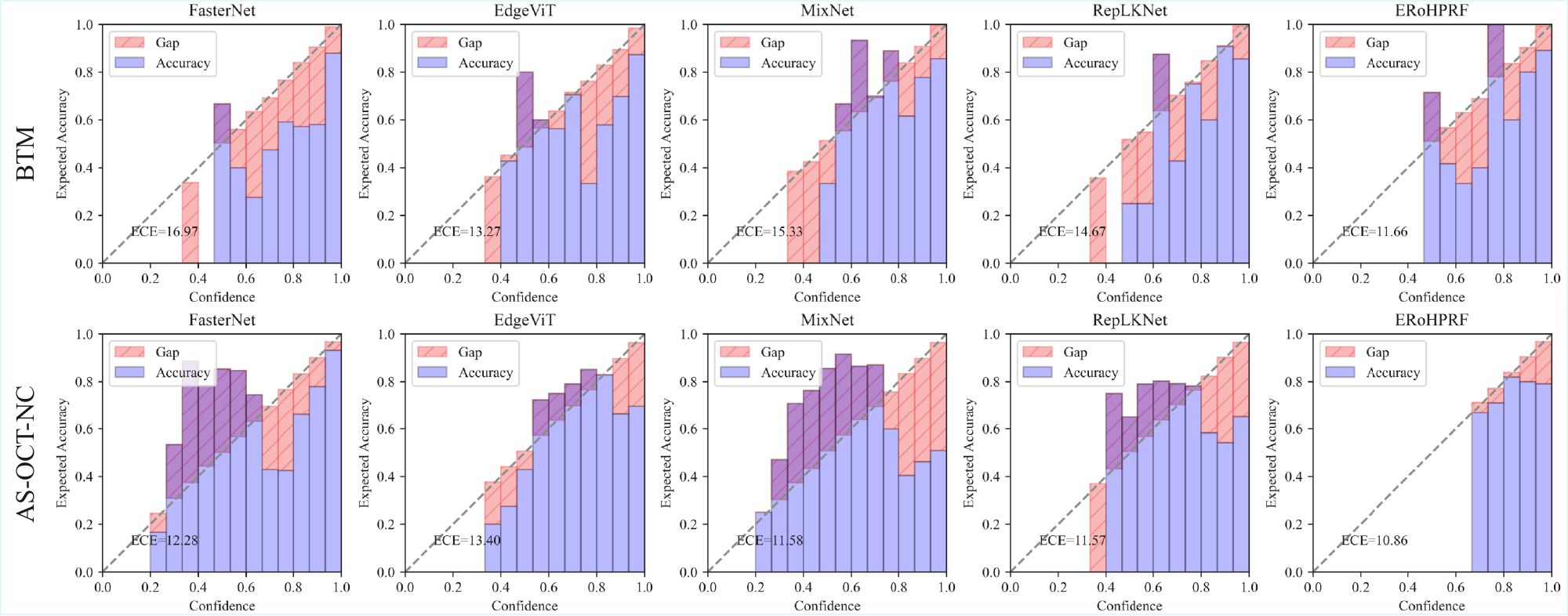}
 \caption{The reliability diagrams of our ERoHPRF, three advanced DNNs, and one structural reparameterization method on the BTM dataset and the AS-OCT-NC dataset.}
\label{ECE_bc}
\end{figure*}

\begin{figure*}
    \centering
    \includegraphics[width=0.95\linewidth]{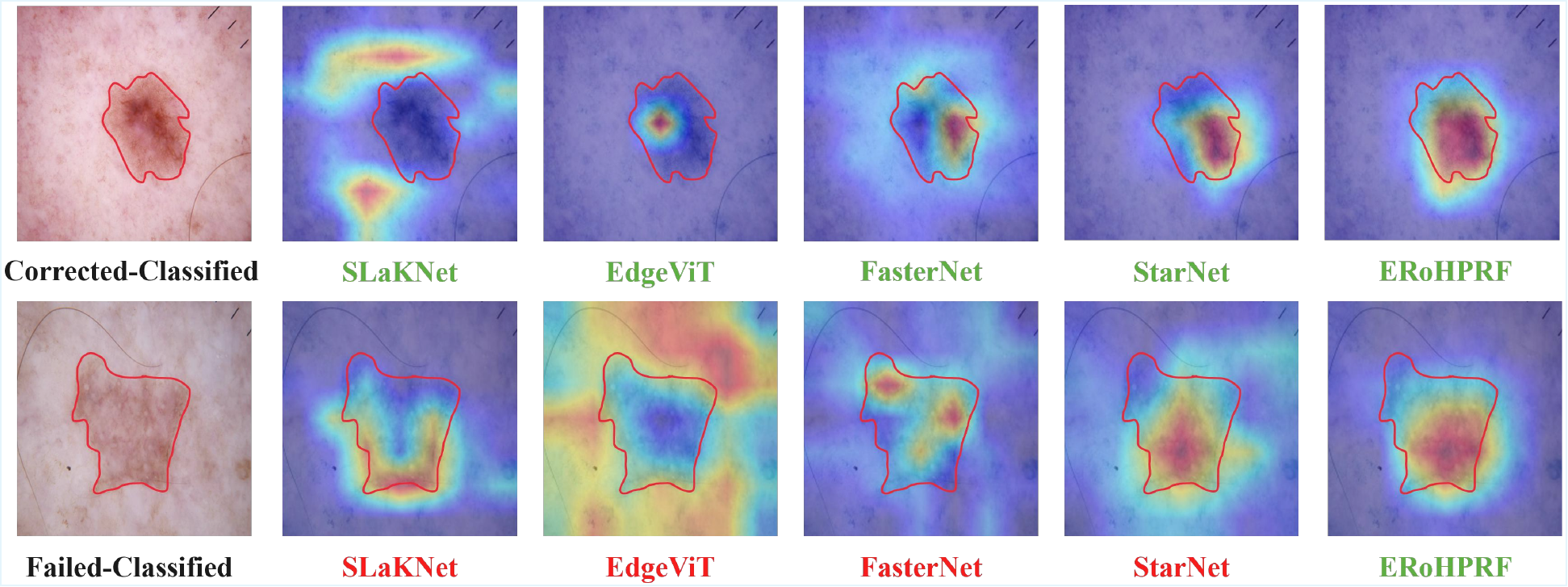}
    \caption{Heat feature maps of failed-classified examples and corrected-classified examples produced by our ERoHPRF and other representative DNNs through Grad-CAM techniques based on the ISIC2018 dataset. The green name indicates corrected-classified, and the red name indicates failed-classified. The lesions are circled in red.}
   \label{heat_isic}
\end{figure*}

\begin{figure*}
    \centering
    \includegraphics[width=0.95\linewidth]{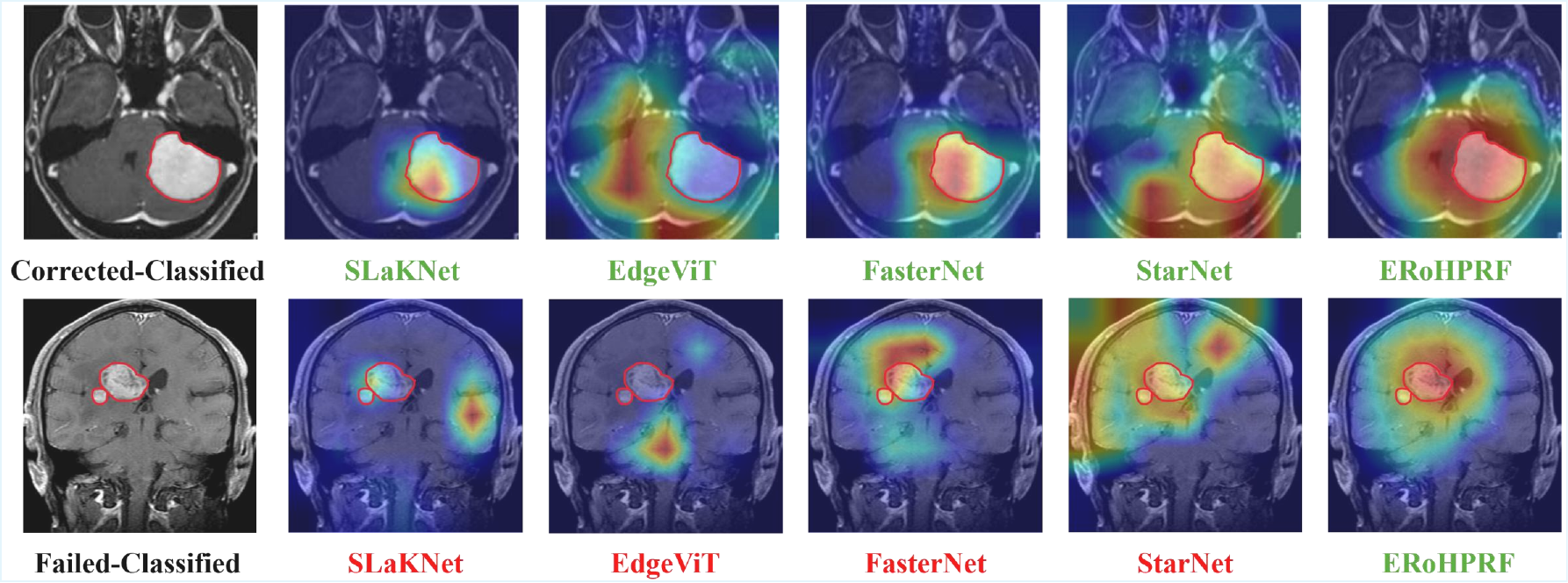}
    \caption{Heat feature maps of failed-classified examples and corrected-classified examples produced by our ERoHPRF and other representative DNNs through Grad-CAM techniques based on the BTM dataset. The green name indicates corrected-classified, and the red name indicates failed-classified. The lesions are circled in red.}
   \label{heat_btm}
\end{figure*}

We further analyze the decision-making fairness of ERoHPRF and the other four representative methods through ROC curves and reliability diagrams on four imbalanced/balanced medical image datasets, as presented in Fig.~\ref{AUC}-Fig.~\ref{ECE_bc}. Fig.~\ref{AUC} offers the ROC curves of our ERoHPRF, three advanced DNNs, and one structural reparameterization method. We observe that our ERoHPRF gets the highest AUC values among all methods through all four datasets, demonstrating the effective exploitation of a multi-expert prior of diverse lesion characteristics assessment mode for heterogeneous pyramid RFs design. Fig.~\ref{ECE_ai} presents the reliability diagrams of ECE for our ERoHPRF and other comparable methods on imbalanced APTOS2019 and ISIC2018. It seems that ERoHPRF boosts the decision-making fairness by obtaining a lower ECE than others. Additionally, Fig.~\ref{ECE_bc} presents the reliability diagrams of ECE for ERoHPRF and the other four methods on balanced BTM and AS-OCT-NC, and we obtain similar conclusions, aligning with our expectations. Fig.~\ref{heat_isic} and Fig.~\ref{heat_btm} offer the heat feature maps of failed-classified examples and corrected-classified examples produced by our ERoHPRF and other representative DNNs through Grad-CAM techniques.  It can be seen that our ERoHPRF pays more attention to diverse lesions compared to other DNNs in both failed-classified and corrected-classified examples, explaining why it performs better in medical image classification and fairness.

In general, the above efficiency and fairness analysis manifest that our ERoHPRF keeps a more promising balance than other comparable DNNs in terms of computational overhead and fairness on imbalanced/balanced medical image classification, proving the superiority of our method in mining the merits of different lesion representations from the perspective of heterogeneous pyramid RFs.

\begin{table}
\caption{Performance comparison of our ERoHPRF and four other representative DNNs in terms of medical image classification and fairness under few-shot learning conditions based on the APTOS2019 dataset.}
    \centering
    \resizebox{1.0\linewidth}{!}{
    \begin{tabular}{c|ccc|cccc} 
        \toprule
         \multirow{2}{*}{Method}&\multicolumn{3}{c|}{Classification Metrics}&\multicolumn{4}{c}{Fairness Metrics}\\
         &ACC$\uparrow$ &bACC$\uparrow$ &mF1$\uparrow$ &AUC$\uparrow$ &ECE$\downarrow$ &BS$\downarrow$ &CECE$\downarrow$\\
        \midrule
        VGG13~\cite{2015VGG} &72.91&41.40&41.37&82.34&7.00&36.83&34.04\\
        RepLKNet~\cite{2022RepLKNet} &72.18&39.42&38.30&82.29&7.66&39.64&34.94\\
        EdgeViT~\cite{pan2022edgevits} &74.55&41.91&41.13&82.37&5.64&36.94&33.64\\
        EfficientNet~\cite{2019EfficientNet} &71.46&37.77&34.85&80.40&6.25&39.61&37.10\\
        \textbf{ERoHPRF} &\textbf{75.64}&\textbf{51.14}&\textbf{52.29}&\textbf{82.94}&\textbf{5.27}&\textbf{35.89}&\textbf{27.38}\\
        \bottomrule
    \end{tabular}}
    \label{table:fewshot}
\end{table}

Table~\ref{table:fewshot} presents medical image classification and fairness comparisons of our ERoHPRF and four other representative DNNs based on few-shot learning conditions, where the training data was reduced by 75\%. It can be observed that ERoHPRF achieves better medical image classification and fairness performance than other methods, verifying the generalization of our proposed method.

\subsection{Ablation Study}
In this section, we conduct a series of ablation experiments to investigate the effectiveness of different RF types and settings in heterogeneous pyramid RF bag of our ERoHPRF. 

\textbf{Effects of Different RF Types.} Table~\ref{table:ablation-type} presents the results of ERoHPRF and its variants, where suffix "-S", "-A", and "-R" denote the ERoHPRF with square RF, coordination RFs, and rectangular RFs, respectively. It can be seen that ERoHPRF performs better than others, and each RF type brings helpful performance improvement in classification and fairness. It also proves that various RF types are helpful in capturing different lesion representations by mimicking the multi-expert consultation mode in aggregating the relative roles of different lesion types. Moreover, Table~\ref{table:efficiency_type} presents the efficiency comparisons of different RF types based on APTOS2019. It shows that the computational overhead increases of our ERoHPRF along with the growth of the number of branches. Since ERoHPRF extends heterogeneous pyramid RFs to one extensive RF in the single square convolution via an expert-like reparameterization operator in the inference stage, they have the same computation overhead in the inference stage without the effects of hardware environments.

\begin{table}
    \caption{Performance comparisons of different RF types based on the APTOS2019 dataset.}
    \centering
    \resizebox{1.0\linewidth}{!}{
    \begin{tabular}{c|ccc|cccc} 
        \toprule
         \multirow{3}{*}{RF Type}&\multicolumn{3}{c|}{Classification Metrics}&\multicolumn{4}{c}{Fairness Metrics}\\
         &ACC$\uparrow$ &bACC$\uparrow$ &mF1$\uparrow$ &AUC$\uparrow$ &ECE$\downarrow$ &BS$\downarrow$ &CECE$\downarrow$\\
        \midrule
        \textbf{ERoHPRF} &\textbf{84.00}&\textbf{63.41}&\textbf{66.13}&86.87&\textbf{3.01}&\textbf{24.03}&\textbf{20.22}\\
        -S &81.82&60.10&62.79&89.55&3.28&26.23&21.06\\
        -A &79.27&49.40&49.88&87.62&3.35&29.58&31.82\\
        -R &80.73&60.80&62.09&\textbf{89.65}&3.30&27.73&20.75\\
        -SA &82.55&62.07&64.58&87.77&3.60&25.57&21.19\\
        -SR &82.73&62.22&63.93&89.43&4.85&24.54&21.46\\
        -AR &82.18&59.43&62.32&85.90&5.94&26.83&24.83\\
        \bottomrule
    \end{tabular}}
    \label{table:ablation-type}
\end{table}

\begin{table}
\caption{Efficiency comparisons of different RF types based on the APTOS2019 dataset.}
    \centering
    \resizebox{1.0\linewidth}{!}{
    \begin{tabular}{c|ccc|ccc} 
        \toprule
         \multirow{2}{*}{Method}&\multicolumn{3}{c}{Training (Server)}&\multicolumn{3}{c}{Inference (Edge)}\\
         &Params$\downarrow$ &MACs$\downarrow$ &FPS$\uparrow$ &Params$\downarrow$ &MACs$\downarrow$ &FPS$\uparrow$\\
        \midrule
        \textbf{ERoHPRF} &3.929M&1.226G&37&\textbf{2.509M}&\textbf{0.534G}&\textbf{27.0}\\
        -S &2.787M&0.673G&\textbf{98}&2.509M&0.534G&26.8\\
        -A &\textbf{2.451M}&\textbf{0.549G}&69&2.509M&0.534G&26.8\\
        -R &2.994M&0.778G&69&2.509M&0.534G&26.7\\
        -SA &3.087M&0.835G&52&2.509M&0.534G&26.7\\
        -SR &3.629M&1.064G&52&2.509M&0.534G&26.9\\
        -AR &3.294M&0.940G&42&2.509M&0.534G&26.7\\
        \hline
    \end{tabular}}
    \label{table:efficiency_type}
\end{table}

\begin{figure}
\centering
\includegraphics[width=0.85\linewidth]{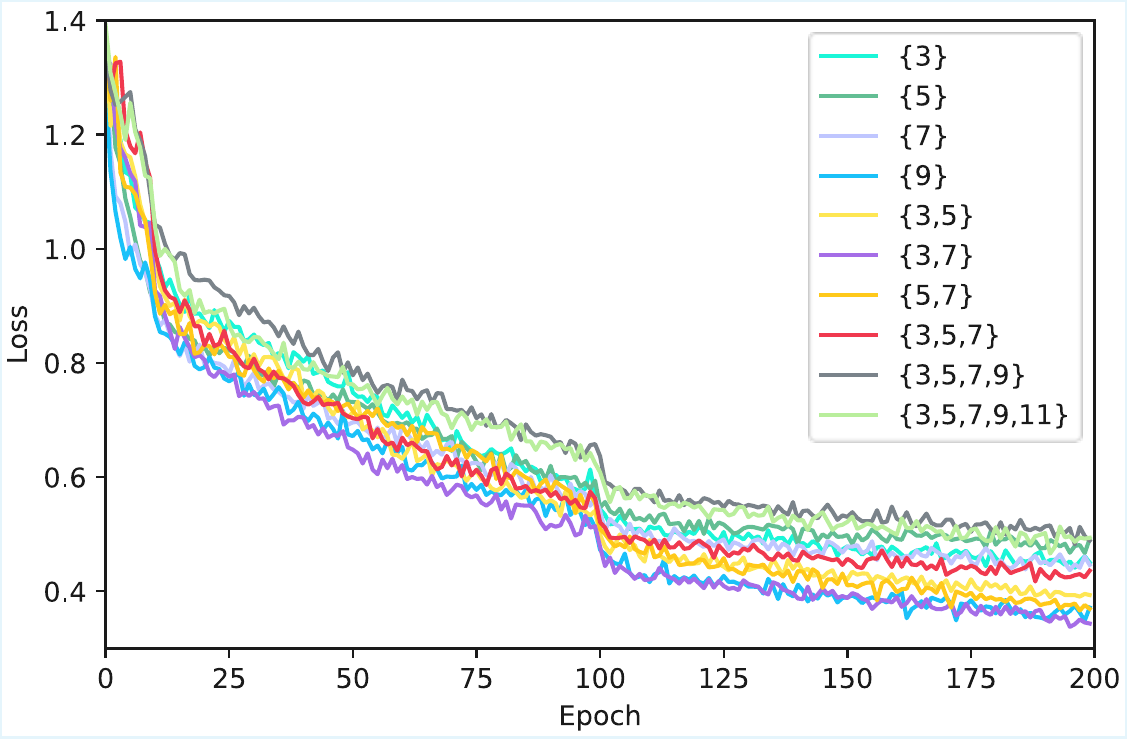}
 \caption{Training loss curves of different expert-like pyramid RF settings based on the APTOS2019 dataset.}
\label{ablation_trainingloss}
\end{figure}

\begin{figure}
\centering
\includegraphics[width=0.85\linewidth]{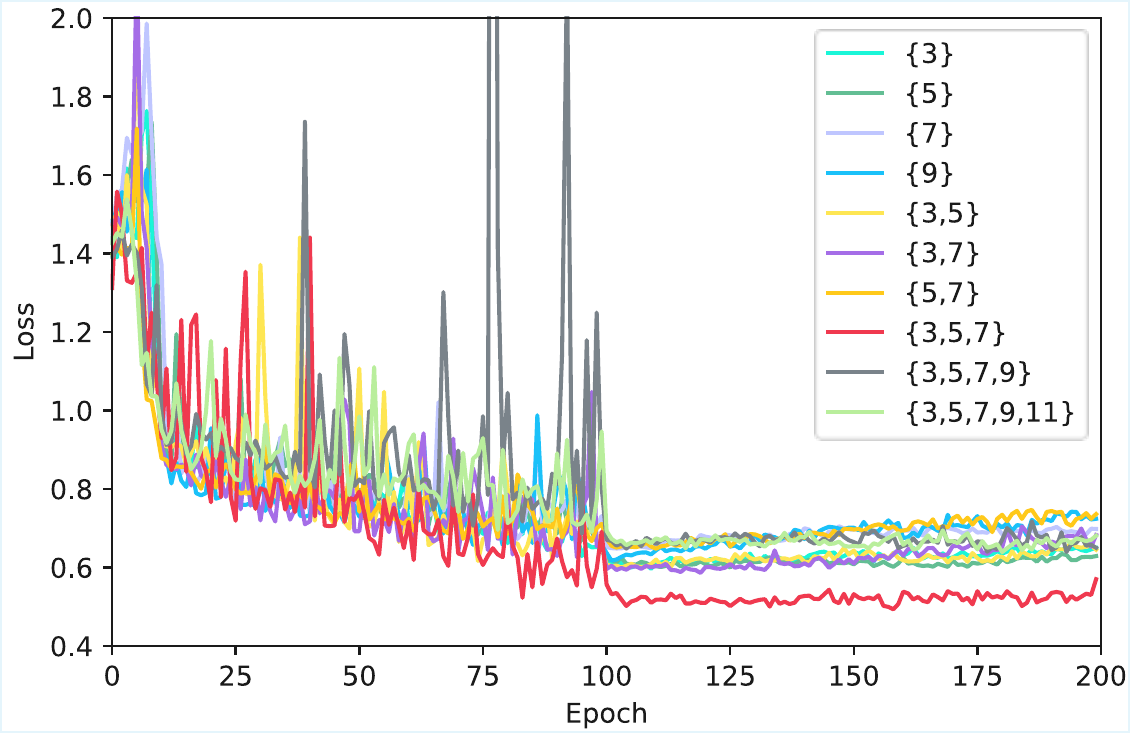}
 \caption{Validation loss curves of different expert-like pyramid RF settings based on the APTOS2019 dataset.}
\label{ablation_validloss}
\end{figure}

\textbf{Effects of Expert-Like Pyramid RF Settings.} Table \ref{table:ablation-size} compares ERoHPRF with different expert-like pyramid RF settings. For example, $\{3\}$ denotes the ERoHPRF with only RF of 3, and $\{5,7\}$ represents the ERoHPRF with RFs of 5 and 7. It is worth noting that larger RFs degrade the performance of ERoHPRF; this might be because larger RFs introduce redundant feature representation information. Thus, it is challenging to set a proper expert-like pyramid RFs, and ERoHPRF with pyramid RFs $\{3, 5, 7\}$ performs better than other RF settings. Fig.~\ref{ablation_trainingloss} and Fig.~\ref{ablation_validloss} offer the training loss and validation loss of different expert-like pyramid RFs settings. We can observe that: 1) small RFs are easy to converge, but it is difficult to capture different lesion representations. 2) Large RFs are hard to converge and have limitations in learning different lesion representations. Besides, Table~\ref{table:efficiency_rf} offers the efficiency comparisons of different expert-like pyramid RF settings based on APTOS2019. Since ERoHPRF utilizes the ELSR in the inference stage, the computation overhead of variants is determined by the largest RF. The computation overhead in training and inference shows a trend of increasing as the RF gets larger. Through comparison, our ERoHPRF with expert-like pyramid RFs $\{3, 5, 7\}$ maintains a better trade-off in RF sizes, efficiency, and fair medical image classification performance.

\begin{table}
    \caption{Performance comparisons of different expert-like pyramid RF settings based on the APTOS2019 dataset.}
    \centering
    \resizebox{1.0\linewidth}{!}{
    \begin{tabular}{c|ccc|cccc} 
        \toprule
         \multirow{2}{*}{RF Setting}&\multicolumn{3}{c|}{Classification Metrics}&\multicolumn{4}{c}{Fairness Metrics}\\
         &ACC$\uparrow$ &bACC$\uparrow$ &mF1$\uparrow$ &AUC$\uparrow$ &ECE$\downarrow$ &BS$\downarrow$ &CECE$\downarrow$\\
        \midrule
        \{3\} &80.55&56.58&58.53&87.43&4.58&26.74&23.70\\
        \{5\} &81.64&58.74&61.80&\textbf{89.44}&4.28&26.40&21.39\\
        \{7\} &81.46&57.42&60.08&88.97&3.95&26.68&26.59\\
        \{9\} &80.91&57.21&60.29&85.66&5.06&26.85&24.65\\
        \{3, 5\} &82.91&59.76&61.91&86.43&3.92&25.37&21.48\\
        \{3, 7\} &82.36&60.80&63.16&86.00&3.23&24.58&20.68\\
        \{5, 7\} &81.09&56.53&58.26&87.13&6.13&26.38&27.18\\
        \textbf{\{3, 5, 7\}} &\textbf{84.00}&\textbf{63.41}&\textbf{66.13}&86.87&3.01&\textbf{24.03}&\textbf{20.22}\\
        \{3, 5, 7, 9\} &81.64&57.84&61.26&87.79&\textbf{2.99}&25.52&21.86\\
        \{3, 5, 7, 9, 11\} &79.09&56.74&58.75&84.57&3.28&28.67&22.39\\
        \bottomrule
    \end{tabular}}
    \label{table:ablation-size}
\end{table}

\begin{table}
\caption{Efficiency comparisons of different expert-like pyramid RF settings based on the APTOS2019 dataset.}
    \centering
    \resizebox{1.0\linewidth}{!}{
    \begin{tabular}{c|ccc|ccc} 
        \toprule
         \multirow{2}{*}{Method}&\multicolumn{3}{c}{Training (Server)}&\multicolumn{3}{c}{Inference (Edge)}\\
         &Params$\downarrow$ &MACs$\downarrow$ &FPS$\uparrow$ &Params$\downarrow$ &MACs$\downarrow$ &FPS$\uparrow$\\
        \midrule
        \{3\} &\textbf{2.373M}&\textbf{0.510G}&80&\textbf{2.223M}&\textbf{0.414G}&27.1\\
        \{5\} &2.687M&0.643G&80&2.337M&0.462G&26.6\\
        \{7\} &3.172M&0.847G&80&2.509M&0.534G&26.9\\
        \{9\} &3.829M&1.124G&\textbf{82}&2.737M&0.631G&26.3\\
        \{3, 5\} &2.908M&0.766G&49&2.337M&0.462G&\textbf{27.1}\\
        \{3, 7\} &3.393M&0.970G&49&2.509M&0.534G&27.0\\
        \{5, 7\} &3.707M&1.103G&49&2.509M&0.534G&27.0\\
        \textbf{\{3, 5, 7\}} &3.929M&1.226G&37&2.509M&0.534G&27.0\\
        \{3, 5, 7, 9\} &5.606M&1.962G&27&2.737M&0.631G&26.4\\
        \{3, 5, 7, 9, 11\} &8.110M&3.048G&22&3.022M&0.751G&26.0\\
        \hline
    \end{tabular}}
    \label{table:efficiency_rf}
\end{table}

\begin{figure*}
    \centering
    \includegraphics[width=0.9\linewidth]{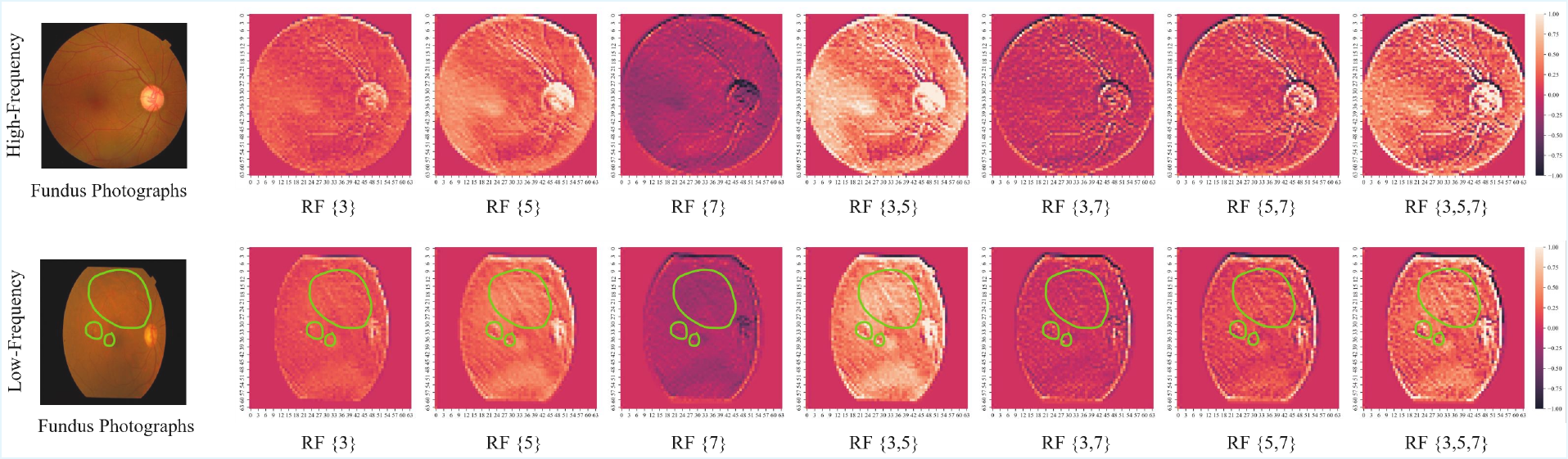}
    \caption{Visualizations of intermediate feature maps produced by HPRFB in the ERoHPRF with different RFs and combinations (\{3\}, \{5\}, \{7\}, \{3, 5\}, \{3, 7\}, \{5, 7\}, and \{3, 5, 7\}). We take two representative images of high-frequency (Normal) and low-frequency (Severe) classes from the APTOS2019 dataset as examples. The lesions are circled in green.}
   \label{visual_APTOS}
\end{figure*}

\begin{figure*}
    \centering
    \includegraphics[width=0.9\linewidth]{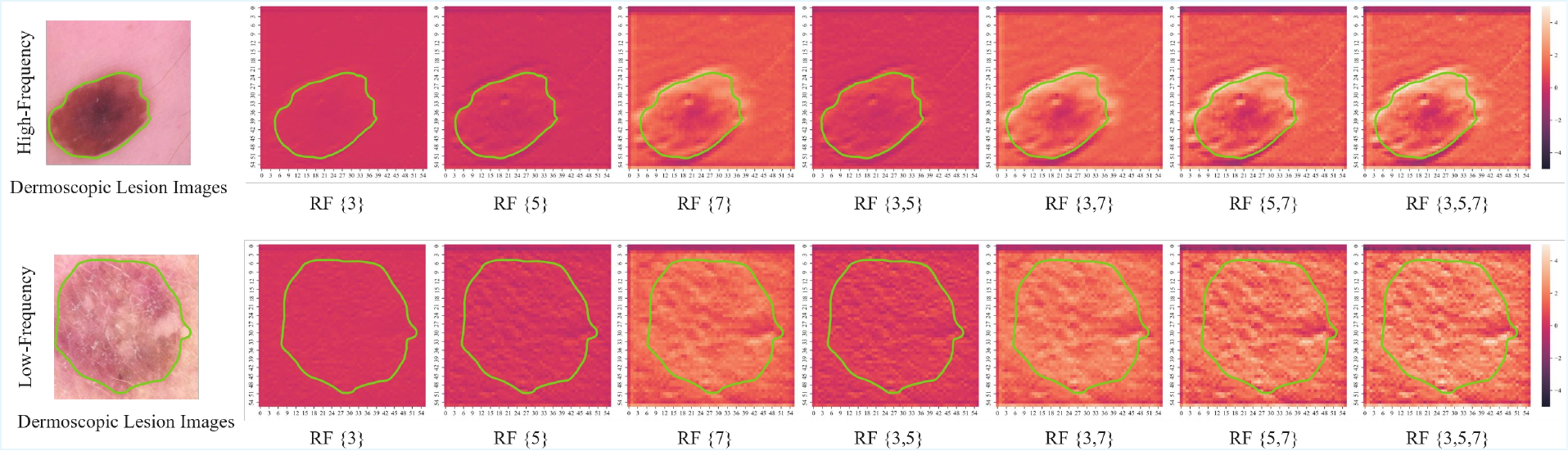}
    \caption{Visualizations of intermediate feature maps produced by HPRFB in the ERoHPRF with different RFs and combinations (\{3\}, \{5\}, \{7\}, \{3, 5\}, \{3, 7\}, \{5, 7\}, and \{3, 5, 7\}). We take two representative images of high-frequency (Nevus) and low-frequency (Dermatofibroma) classes from the ISIC2018 dataset as examples. The lesions are circled in green.}
   \label{visual_ISIC}
\end{figure*}

\begin{figure*}
\centering
\includegraphics[width=0.9\linewidth]{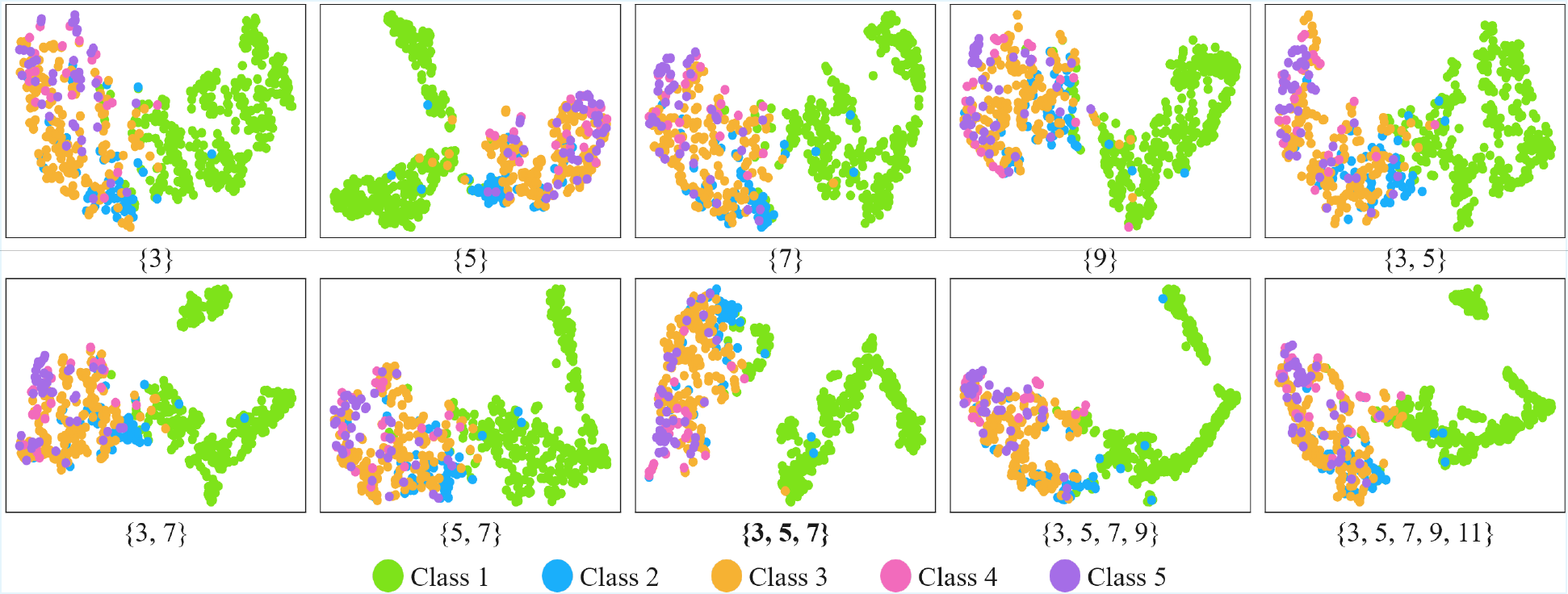}
\caption{t-SNE visualization comparisons of different RF settings in ERoHPRF under the APTOS2019 dataset.}
   \label{visual_tsne}
\end{figure*}

\begin{figure}[!t]
\centering
\includegraphics[width=0.85\linewidth]{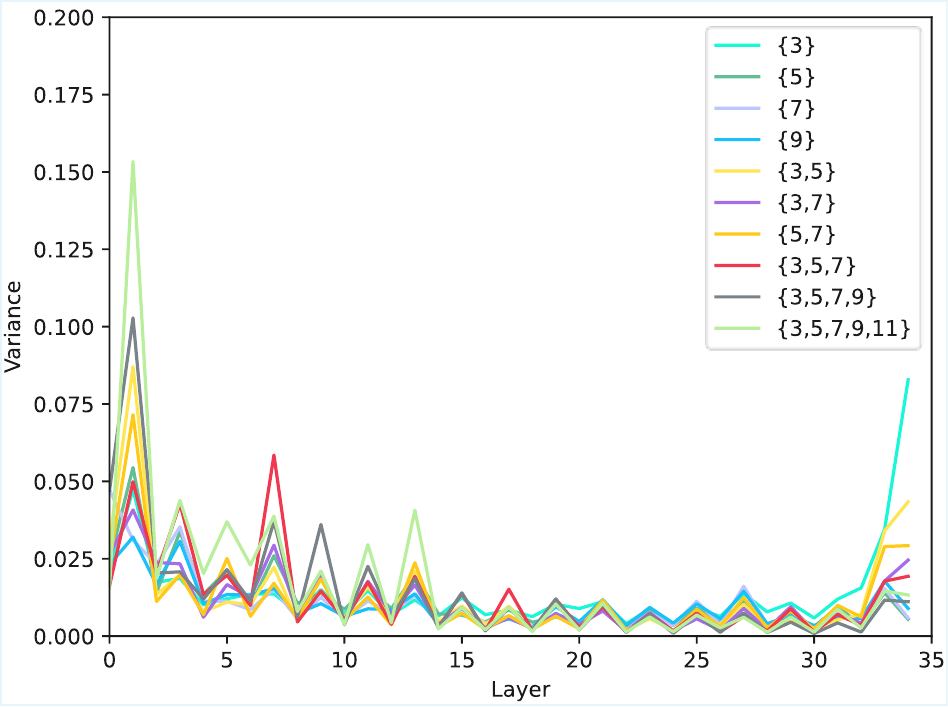}
 \caption{Activation variance curves of different expert-like pyramid RF settings for heterogeneous pyramid RF bag in ERoHPRF based on the APTOS2019 dataset.}
\label{activation}
\end{figure}

\textbf{Feature Representation Visualization of Expert-Like Pyramid RFs and Heterogeneous RFs.} To better understand the internal behaviors of expert-like pyramid RFs and heterogeneous RFs in capturing diverse lesion characteristics concurrently, Fig.~\ref{visual_APTOS} and Fig.~\ref{visual_ISIC} offer intermediate feature map visualizations produced by HPRFB in the ERoHPRF with different RFs and combinations (\{3\}, \{5\}, \{7\}, \{3, 5\}, \{3, 7\}, \{5, 7\}, and \{3, 5, 7\}). Fig.~\ref{visual_tsne} presents the t-SNE visualizations of the learned feature representations of different RFs and their corresponding combinations for HPRFB in the ERoHPRF. We observe that our ERoHPRF with RF setting \{3, 5, 7\} performs better in maximizing the inter-class distance and minimizing the intra-class distance, manifesting that heterogeneous pyramid RF bag are capable of guiding CNNs to pay attention to diverse lesion representations, such as boundary, tiny, coordination, small, and salient. Fig.~\ref{activation} presents the activation variance visualizations of different RF settings in the ERoHPRF. It is seen that variance values at the shallow layers are larger than those at the deep layers, verifying that our ERoHPRF captures diverse lesion representations effectively by adjusting the relative significances of heterogeneous pyramid RFs, supporting our hypothesis in mimicking the multi-expert consultation mode.

\section{Conclusion and Future Work}
Inspired by the multi-expert consultation mode, this paper introduces a new concept of ERoHPRF for efficient CNN design, which incorporates the multi-expert prior of diverse lesion characteristics into the efficient convolution operator from the aspect of heterogeneous pyramid RFs. The extensive experiments on balanced/imbalanced medical image datasets manifest that our method maintains a better trade-off than SOTA structural reparameterization methods, efficient CNNs, and advanced transformers in terms of classification, fairness, and computation overhead. However, our method still has some limitations: 1) The Theoretical analysis of ERoHPRF is not enough. 2) ERoHPRF can not assign different weights to heterogeneous pyramid RFs dynamically. 3) The fairness of ERoHPRF needs further improvements, as well as other modality information, e.g., demographics, which might boost the fairness of our method. 

In the future, we plan to investigate the superiority and generalization of our method in other medical image analysis tasks, such as medical image segmentation or detection, by treating it as a plug-and-play convolution module.

\bibliographystyle{IEEEtran}
\bibliography{tmi}

\end{document}